\documentclass[journal]{IEEEtran}
\usepackage[pdftex]{graphicx}
\ifCLASSINFOpdf
\else
\fi
\usepackage{amsmath}
\usepackage{subcaption}
\usepackage{fdsymbol}
\usepackage[flushleft]{threeparttable}
\usepackage{xcolor}
\usepackage{soul}
\usepackage[acronym,shortcuts,nomain,xindy,toc]{glossaries}
\newacronym[]{GPS}{GPS}{Global Positioning System}
\newacronym[]{VUT}{VUT}{Vehicle Under Test}
\newacronym[]{B2P}{B2P}{Bumper to Pedestrian}
\newacronym[]{GVT}{GVT}{Global Vehicle Target}
\newacronym[]{TTC}{TTC}{Time To Collision}
\newacronym[]{AD}{AD}{Automated Driving}
\newacronym[]{VRU}{VRU}{Vulnerable Road Users}

\defglsdisplayfirst[\acronymtype]{\textit{#1}}

\begin{document}
%
\title{Testing predictive automated driving systems: lessons learned and future recommendations}
%
%
%

\author{Rubén~Izquierdo~Gonzalo, Carlota~Salinas~Maldonado, Javier~Alonso~Ruiz, Ignacio~Parra~Alonso, David~Fernández~Llorca and Miguel~Á. Sotelo 

\thanks{R. Izquierdo, C. Salinas, J. Alonso, I. Parra and M.A. Sotelo are with the Computer Engineering Department, University of Alcalá, Alcalá de Henares, Madrid, Spain e-mail: \{ruben.izquierdo, carlota.salinas, javier.alonso, ignacio.parra,  miguel.sotelo\}@uah.es.}
\thanks{D. Fernández~Llorca is with the Joint Research Centre, European Commission, Sevilla, Spain e-mail: david.fernandez-llorca@ec.europa.eu.}
\thanks{Manuscript received November 5, 2021; revised MMMM DD, YYYY.}
}

\maketitle
\begin{abstract}
Conventional vehicles are certified through classical approaches, where different physical certification tests are set up on test tracks to assess required safety levels. These approaches are well suited for vehicles with limited complexity and limited interactions with other entities as last-second resources. However, these approaches do not allow to evaluate safety with real behaviors for critical and edge cases, nor to evaluate the ability to anticipate them in the mid or long term. This is particularly relevant for automated and autonomous driving functions that make use of advanced predictive systems to anticipate future actions and motions to be considered in the path planning layer. In this paper, we present and analyze the results of physical tests on proving grounds of several predictive systems in automated driving functions developed within the framework of the BRAVE project. Based on our experience in testing predictive automated driving functions, we identify the main limitations of current physical testing approaches when dealing with predictive systems, analyze the main challenges ahead, and provide a set of practical actions and recommendations to consider in future physical testing procedures for automated and autonomous driving functions.

    
    
    
    
    
    
    
    
\end{abstract}


%
\IEEEpeerreviewmaketitle


\section{Introduction}
%
%
%
%
\IEEEPARstart{I}{n} order to place a vehicle on the market, car manufacturers need prior authorization, granted by the competent authority, after proving that the vehicle complies with all applicable regulatory standards and safety certification requirements. Whether through vehicle \emph{type approval} or \emph{self-certification} approaches, Original Equipment Manufacturers (OEMs) must pass stringent certification processes to validate a component, a system, or the entire vehicle \cite{Martins2010}.

Conventional vehicles are certified through classical approaches, where different physical certification tests are set up on test tracks or test benches to assess the required safety level using various performance criteria. These approaches are well suited for components, systems, and vehicles with limited complexity and limited interactions with other entities (e.g, braking tests). However, as the complexity of systems increases (e.g., Electronic Stability Control), 
classical approaches cannot address all relevant safety areas due to two main reasons. First, the large number of safety-related systems (including multiple electrical and electronic systems) which increases risks from systematic failures and random hardware failures. This is reasonably well addressed by the existing functional safety and Automotive Safety Integrity Levels (ASIL) requirements in the automotive industry (e.g., ISO 26262). Second, the enormous variability of possible multi-agent scenarios, which, on the one hand, implies the need for a formal safety model \cite{Shashua2018}, and on other hand, has led to the introduction of simulation-based safety-oriented audits, as a way to complement physical vehicle testing \cite{Lutz2017}.

With the introduction of assisted (SAE Levels 1 and 2), automated (SAE Level 3), and autonomous (SAE Levels 4 and 5) driving systems \cite{Schram2019}, \cite{Llorca2021}, the overall complexity increases in terms of the number of software functions, variants of multi-agent scenarios and interactions, and potentially affected safety areas \cite{Edwards2016}. The complexity of these systems, and therefore the difficulty to test them, increases with the level of automation, being particularly important the step from SAE Level 3 to 4 since the automated driving system must be able to reach a \emph{minimal risk condition} within its \emph{Operational Design Domain} (ODD) without user/passenger intervention \cite{SAE2021}.

New innovative testing approaches, including procedures of different nature, are needed for future vehicle safety regulatory frameworks and for assessments under current exemption procedures \cite{Galassi2020}. New online/in-service safety monitoring and verification mechanisms \cite{Althoff2014} that act after the market deployment of automated driving systems \cite{Edwards2016} are also needed as a way of reducing the need to test all possible combinations at the time of type-approval. Several national and international regulatory and standardization initiatives and projects are already underway to tackle all these problems \cite{Baldini2020}. 


One of the most solid regulatory proposals is being developed by the Working Party on Automated/Autonomous and Connected Vehicles (GRVA) of the UNECE World Forum for Harmonization of Vehicle Regulations (WP.29). It is based on three pillars that must be assessed together \cite{GRVA2019}. First, \emph{audit and assessment} which includes the use of simulation to cover all types of scenarios, but especially edge case scenarios difficult to occur in real-world traffic. Second, \emph{physical certification tests} to assess critical scenarios, performed in controlled environments on test tracks (closed-roads), and involving sophisticated equipment such as lightweight global vehicle \cite{EuroNCAP-GVT}, articulated pedestrian \cite{ACEA-APT} and bicyclist \cite{ACEA-BT} targets.  And finally, \emph{real-world test drive}, which is devised as a “driving license test” for automated driving systems to assess the overall capabilities and behavior of the vehicle in non-simulated traffic on public or open roads. This approach has been the one adopted by UN to regulate the approval of Advanced Emergency Braking Systems (AEBS) \cite{UN-AEBS2020} and, more recently, Automated Lane Keeping Systems (ALKS) \cite{UN-ALKS2021}. These regulations have recently been integrated in countries such as Japan and Germany, enabling the commercialization of the first SAE Level 3 automated driving systems by two different car manufacturers \cite{Honda2021}, \cite{MB2021}.

A similar approach was provided by the PEGASUS project \cite{PEGASUS}, including laboratory, simulation, testing site tests, and field tests, with particular emphasis on the definition of use cases and test scenarios. In another project, ENABLE-S3 \cite{ENABLE-S3}, the goal was to reduce testing efforts of traditional road testing by focusing on virtualization. One of the main contributions was the use of semi-virtual systems, such as the DrivingCube \cite{DrivingCube} which combines both simulation and ready-to-drive vehicles on a chassis dynamometer and on a power-train testbed. This approach can be considered as an intermediate step between pure simulation-based verification and physical certification testing, or as a subfield of simulation using vehicle-in-the-loop (VIL).

\begin{table*}[t]
  \begin{threeparttable}
  \centering
    \caption{Main features of the different testing approaches.}
  \begin{tabular}{|l|c|c|c|c|c|c|c|}
  \hline
    \textbf{Approaches} & \textbf{Control./Repeat.} & \textbf{Scalability} & \textbf{Variability} & \textbf{Fidelity} & \textbf{Efficiency} &\textbf{Test Overfitting \tnote{a}} &\textbf{Real Behaviors \tnote{b}} \\ \hline
    Simulation & $\medblackstar \medblackstar \medblackstar \medblackstar \medblackstar$  & $\medblackstar \medblackstar \medblackstar \medblackstar \medblackstar$ & $\medblackstar \medblackstar \medblackstar \medblackstar \medblackstar$ & $\medblackstar \medwhitestar \medwhitestar \medwhitestar \medwhitestar$ & $\medblackstar \medblackstar \medblackstar \medblackstar \medblackstar$ & $\medblackstar \medblackstar \medblackstar \medwhitestar \medwhitestar$ & $\medwhitestar \medwhitestar \medwhitestar \medwhitestar \medwhitestar$ \\ \hline
    VIL Simulation & $\medblackstar \medblackstar \medblackstar \medblackstar \medblackstar$ & $\medblackstar \medblackstar \medblackstar \medblackstar \medwhitestar$ & $\medblackstar \medblackstar \medblackstar \medblackstar \medwhitestar$ & $\medblackstar \medblackstar \medwhitestar \medwhitestar \medwhitestar$ & $\medblackstar \medblackstar \medblackstar \medwhitestar \medwhitestar$ & $\medblackstar \medblackstar \medblackstar \medblackstar \medwhitestar$ & $\medwhitestar \medwhitestar \medwhitestar \medwhitestar \medwhitestar$ \\ \hline
    Physical track & $\medblackstar \medblackstar \medblackstar \medblackstar \medwhitestar$ & $\medblackstar \medblackstar \medblackstar \medwhitestar \medwhitestar$ & $\medblackstar \medwhitestar \medwhitestar \medwhitestar \medwhitestar$ & $\medblackstar \medblackstar \medblackstar \medwhitestar \medwhitestar$ & $\medblackstar \medblackstar \medwhitestar \medwhitestar \medwhitestar$ & $\medblackstar \medwhitestar \medwhitestar \medwhitestar \medwhitestar$ & $\medwhitestar \medwhitestar \medwhitestar \medwhitestar \medwhitestar$\\ \hline
    Open-roads & $\medblackstar \medwhitestar \medwhitestar \medwhitestar \medwhitestar$ & $\medblackstar \medwhitestar \medwhitestar \medwhitestar \medwhitestar$ & $\medblackstar \medblackstar \medwhitestar \medwhitestar \medwhitestar$ & $\medblackstar \medblackstar \medblackstar \medblackstar \medblackstar$ &  $\medblackstar \medblackstar \medblackstar \medwhitestar \medwhitestar$ & $\medblackstar \medblackstar \medblackstar \medblackstar \medwhitestar$ & 
    $\medblackstar \medblackstar \medblackstar \medblackstar \medblackstar$ \\
    \hline
  \end{tabular}
    \begin{tablenotes}
            \item[a] \emph{Test overfitting refers to the degree to which the systems can be optimized on specific test scenarios. A high score means a low probability of overfitting.}
            \item[b] \emph{Real behaviors refers to the degree to which the test method can include actual behaviors of other agents (pedestrians, cyclists, other drivers, etc.).}            
        \end{tablenotes}  
  \label{tab:1}
\end{threeparttable}  
\end{table*}

The three approaches mentioned (i.e, simulation - including VIL simulation-, physical and real-world testing) have strengths and weaknesses \cite{Thorn2018}, which is why it is important to implement them holistically \cite{GRVA2019}. In Table \ref{tab:1} we illustrate the advantages and disadvantages of all testing approaches. As can be observed, the methods are somehow complementary. For example, although simulation-based testing allows full controllability, repeatability, and variability in a very efficient way, they exhibit very low fidelity and lack real-world behaviors. We can increase fidelity at the cost of increasing complexity and thus decreasing efficiency, from VIL simulation to physical testing on closed tracks. But the absence of real behaviors remains a problem, which can only be partially compensated by testing in real traffic conditions (open-roads). 

Another relevant variable refers to the degree to which the driving functions can be optimized on the specific scenarios, which can be seen as a shortcut by OEMs to overfit the performance of their systems to the test scenarios. This has a negative impact on the possible fidelity of the tests, while the performance of the systems in real traffic remains unknown. In general, if the simulation conditions are known a priori, or the physical test conditions in closed tracks, or the proving grounds or the test area in real traffic, all test methods are potentially subject to overfitting. Still, there are some differences. For example, on the one hand, the uncontrolled conditions of open road testing make this method less prone to overfitting. On the other hand, the low variability and the strict control and repeatability conditions of the scenarios in the physical certification on closed roads are favorable conditions for the optimization of the systems to the proposed scenarios.

\begin{figure}[t]
\centering
    \includegraphics[width=0.9\linewidth]{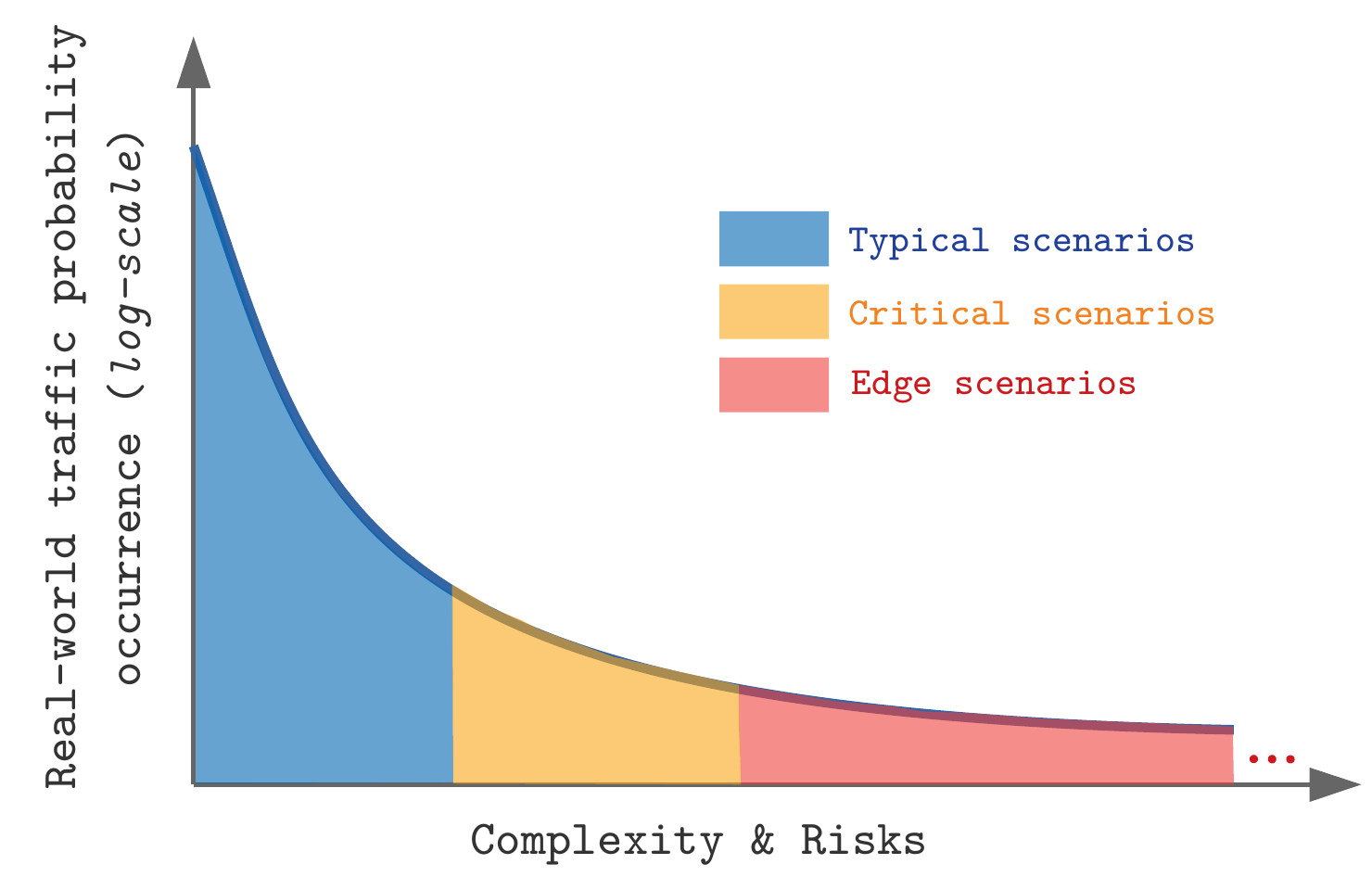}
    \caption{Types of scenarios, probability of occurrence in real-world traffic (log-scale), complexity, and risk. Long-tail distribution. We refer to \cite{GRVA2019} for more details on the scenarios.}
    \label{fig:scenarios}
\end{figure} 

\begin{table}[t]
  \centering
    \caption{Distribution of scenarios by testing approach.}
  \begin{tabular}{|l|c|c|c|}
    \hline
    \textbf{Approaches} & \textbf{Typical} & \textbf{Critical} & \textbf{Edge} \\ \hline
    Simulation & $\checkmark$ & $\checkmark$ & $\checkmark$ \\ \hline
    VIL Simulation & $\checkmark$ & $\checkmark$ & $\checkmark$ \\ \hline
    Physical track & & $\checkmark$ & \\ \hline
    Open-roads & $\checkmark$ &  & \\ \hline    
  \end{tabular}
\label{tab:2}
  \vspace{-5mm}
\end{table}

A closer look at the different methods reveals that complementarity is limited, as the scenarios addressed by each approach are of a different nature. In Fig. \ref{fig:scenarios} we illustrate three different types of scenarios (typical, critical, and edge) referring to their probability of occurrence with respect to the degree of complexity and potential risks. As can be observed the distribution of scenarios follows a long-tail distribution, which requires significant scale \cite{Shashua2018} to discover and properly handle the long tail of rare events \cite{Jain2021}. In Table \ref{tab:2}, the type of scenarios that can be addressed for each testing method is depicted. As can be inferred, high fidelity is only achievable for typical scenarios, with higher uncertainty for critical and edge scenarios. In addition, current testing approaches do not allow to assess safety with real behaviors for critical and edge cases. This is particularly relevant for automated and autonomous driving functions that make use of predictive perception, i.e., systems that learn and model the behaviors and interactions of traffic agents to anticipate future actions and motions to be considered in the path planning layer. In fact, higher levels of automation (autonomous) are expected to be achieved thanks to predictive capabilities \cite{Llorca2021}. These predictive systems are expected to enable autonomous driving to become more like manual driving, increasing safety margins, reducing risks, and providing smoother and more acceptable motion trajectories.  All these factors have a direct effect on the perceived safety, risk and trust of users, which are directly linked to user acceptance \cite{Kaye2021}.

Incorporating real behaviors in critical and edge scenarios, both in simulation environments and in physical tests on tracks, is not straightforward. However, the advantages in safety and comfort \cite{Elbanhawi2015} that predictive autonomous driving brings require new efforts to improve test methods for future certification processes. In an attempt to take the first steps, in this paper we present and analyze the results of physical tests on proving grounds of several predictive systems in automated driving functions developed within the framework of the BRAVE project \cite{BRAVE}. A number of use cases involving vehicles and Vulnerable Road Users (VRUs) on different scenarios were defined, some of them directly equivalent to the EuroNCAP test protocols for Automatic Emergency Braking (AEB) systems \cite{EuroNCAP-AEB-Car2019}, \cite{EuroNCAP-AEB-VRU2020}. 

We focus our work on physical tests because they have a higher level of maturity (e.g. EuroNCAP test protocols) and offer more room for improvement in terms of variability and overfitting, as well as a better relationship between fidelity and controllability/repeatability \cite{Thorn2018}. These tests are also essential to validate the fidelity (reality gap) of simulation-based methods \cite{GRVA2019}. Based on our experience in testing predictive automated driving functions, we identify the main limitations of current physical testing approaches when dealing with predictive systems, analyze the main challenges ahead and provide a set of practical actions and recommendations to consider in future physical testing procedures for automated and autonomous driving functions. 

From this practical bottom-up approach (from direct empirical evidence in testing to conceptual identification of needs for future certification processes) we aim to contribute to and complement the high-level approaches carried out worldwide to adapt regulatory standards and safety certification requirements for increasingly advanced autonomous driving systems.

The remainder of this paper is as follows. In Section \ref{sec:overviewpred} we briefly summarize the main predictive perception approaches applied in our automated vehicle. Section \ref{sec:experiments} provides a detailed description of the use cases, scenarios, and results obtained when testing the predictive systems. In Section \ref{sec:discussion} we discuss current limitations, identify the challenges, and provide recommendations for future safety certification requirements. We finalize establishing the main conclusions in Section \ref{sec:conclusions}. 



\section{Overview of predictive perception systems}
\label{sec:overviewpred}

The experiments were performed using UAH's DRIVERTIVE  vehicle \cite{Drivertive}. DRIVERTIVE is a mechanically automated vehicle that carries multiple sensors for environment perception, such as an HDL-32E LiDAR, 3 RADARs, and multiple cameras. The objective of the experiments was to validate the goodness of predictive systems to outperform conventional last-second reaction systems. These experiments involved different traffic agents, such as vehicles, pedestrians, and cyclists in critical, controlled situations. For self-containment reasons, in this section we briefly introduce the predictive systems used in the experiments.

Regarding the interaction with vehicles, two deep-learning approaches were used for the inference of both intention and trajectories of vehicles. The intention prediction system uses a classification approach with a Resnet50 \cite{resnet50} backbone to classify enhanced images encoded as a single RGB image, including context, interactions, and vehicle motion patterns \cite{ruben_maneuver}. This model estimates the probability of keeping or changing lanes. The trajectory prediction model \cite{ruben_traj} uses radar targets to create a top-view time-evolving map. The most likely future representation of this input map is inferred by a U-net model \cite{u-net}, \cite{Izquierdo2021} trained for this purpose. The output of this model is used as a triggering mechanism to assess critical cut-in maneuvers.

For pedestrians, body and facial keypoints detectors \cite{keypoints} act as the core for prediction systems. Deep learning approaches use body keypoints to anticipate changes in pedestrian motion patterns \cite{raul_pose} and also to predict the intention of crossing from the sidewalk or at a crosswalk \cite{Lorenzo2020}, \cite{Lorenzo2021}. Face key points are of paramount importance for the detection of crossing intention, being eye contact a powerful non-verbal channel of communication often used to express intention to drivers. This eye-contact detection was very useful in improving response time in tests with moving head dummies using the eye-contact signal as a trigger for the breaking response. 

Finally, interaction experiments with bicyclist dummies were conducted using an instance segmentation approach. This method provides object-level detections that allow for radar data fusion. Cyclist path prediction was implemented using a standard approach based on a physical model \cite{Zernetsch2016}. However, in the tests, no actions were expected from the cyclist (e.g., switching dynamics \cite{Kooij2019}) and nothing could be predicted. Interaction with the dummy was limited to maintaining the safety distance, being detected at 100 meters, which provided enough space to ensure a smooth and safe maneuver.
\section{Experimental validation of predictive automated systems}
\label{sec:experiments}

This section provides experimental results derived from physical tests developed on the proving ground at \emph{Union Technique de l'Automobile et du Cycle} (UTAC) facilities \cite{UTAC}. These tests are intended to evaluate autonomous capabilities at standardized EuroNCAP protocol tests as well as custom-defined autonomous tests. The goal of these experiments is to validate and measure the safety and goodness of predictive systems in critical and controlled circumstances. This section begins with a description of the experiments and their correspondence with EuroNCAP tests. For the sake of space, the most relevant configurations are analyzed in detail in two subsections: subsection \ref{subsec:vru} related to \ac{VRU}, i.e., pedestrians and bicyclists interactions, and subsection \ref{subsec:veh} related to vehicle-vehicle interactions.

\subsection{Description of Experiments}
\begin{figure}[ht]
\centering
    \includegraphics[width=0.7\linewidth]{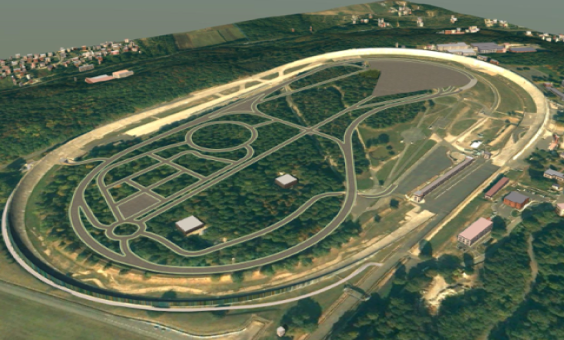}
    \caption{Proving grounds at UTAC's premises.}
    \label{fig:utac_map}
\end{figure}

Several rounds of experimentation were conducted in proving grounds at UTAC (depicted in Fig. \ref{fig:utac_map}), a technology center located at Linas-Monthléry, France. The experiments were carried out in the framework of the BRAVE project \cite{BRAVE} using UAH’s DRIVERTIVE vehicle. During the experiments with the \ac{VUT} the following use cases were tested: Pedestrian (VRU-1 and VRU-2), Cyclist \mbox{(VRU-3)}, Cut-in vehicle (VEH-1) and Intersection (VEH-2). A detailed description of each use case is provided in the next section. All use cases were designed and tested to efficiently assess the performance and the added value of the predictive systems developed in BRAVE.

Some of the tests conducted in this work have a direct equivalence with EuroNCAP use cases. Other tests have been specifically devised and designed to be tested in the BRAVE project, as a means to provide further recommendations to EuroNCAP. Table \ref{table:usecases} provides the equivalence between tests conducted in BRAVE and use cases defined by EuroNCAP.

\begin{table}[ht]
\renewcommand{\arraystretch}{1.2}
    \centering
    \caption{Equivalence between BRAVE and EuroNCAP use cases.}
    \label{table:usecases}
    \begin{tabular}{l|c}
                                    & \textbf{EuroNCAP} \\
    \textbf{BRAVE test}                                 & \textbf{Test Equivalence} \\
    \hline
    \textbf{VRU-1 GRAIL}	                            & \\ 
    Config. 1: crossing on green LED signal             & -- \\
    Config. 2: CPLA-50 deceleration then overtake	    & CPLA-50 day\\
    Config. 3: longitudinal then crossing on green LED  & -- \\
    \hline
    \textbf{VRU-2 PEDESTRIAN}	                        & \\
    Config. 1: CPNA-50. full stop	                    & CPFA-50 \\
    Config. 2: CPNA-O-50, full stop	                    & CPNC-50$^{1}$ \\
    Config. 3: CPNA-25, AES	                            & CPNA-25 day \\
    \hline
    \textbf{VRU-3 CYCLIST}                              & \\	
    Config. 1: CBLA-25, overtake                        & -- \\	
    Config. 2: CBLA-25, deceleration	                & CBLA-50 day \\
    Config. 3: CBLA-25, deceleration then overtake      & -- \\
    \hline
    \textbf{VEH-1 CUT-IN}                                & \\
    Config. 1: Deceleration within central lane         & Cut-in 50/10 \\
    Config. 2: AES onto the left lane                   & -- \\
    \hline
    \textbf{VEH-2 INTERSECTION}                       & \\
    Config. 1: Straight line with crossing target       & -- \\
    Config. 2: Straight line with turning target	    & -- \\
\hline
\multicolumn{2}{@{}p{21pc}@{}}{\it $^1$ The position of the car that causes the occlusion is lateral rather than longitudinal. The dummy corresponds to an adult instead of a child.} \\
    \end{tabular}
\end{table}

As can be observed, the VEH-2 Intersection use case is totally new, having no equivalence in Euro NCAP tests. It has been completely devised and designed in the framework of the BRAVE project in an attempt to provide recommendations to Euro NCAP on how to properly test automated vehicles when dealing with interactions with other vehicles at intersections.

\subsection{Assessment of Vehicle-VRU Interaction}\label{subsec:vru}
Several tests were conducted to assess Vehicle-VRU interactions. For each use case, several configurations were exhaustively tested, as described in Table \ref{table:tabla2}. The schematic description of all the use cases tested is provided as follows.

\begin{itemize}
    \item VRU-1: An external HMI (GRAIL system \cite{GRAIL}) onboard the vehicle is used to interact with pedestrians willing to cross the street. 
    \item VRU-2: Emergency reactions of the VUT are tested in different situations when interacting with pedestrians whose predicted trajectory intersects the VUT estimated trajectory. 
    \item VRU-3: Emergency reactions of the VUT are tested when interacting with cyclists. 
\end{itemize}

\begin{table}[ht]
\renewcommand{\arraystretch}{1.2}
    \centering
    \caption{Description of configurations in VRUs’ use cases.}
    \label{table:tabla2}
    \begin{tabular}{l|l}

    \textbf{VRU}         & \\
    \textbf{Use Case}    & \textbf{Configuration} \\
    \hline
    \textbf{VRU-1}	     & Config. 1: pedestrian crossing at 50\% impact (stop)\\ 
                 	     & Config. 2: pedestrian walking in parallel (not crossing)\\ 
                 	     & Config. 3: pedestrian turns head and crosses\\
    \hline
    \textbf{VRU-2}	     & Config. 1: pedestrian crossing at 50\% impact (stop)\\
                         & Config. 2: occluded pedestrian crossing at 50\% impact (stop)\\
                         & Config. 3: pedestrian crossing at 25\% impact (avoid) \\
    \hline
    \textbf{VRU-3}	     & Config. 1: Overtake \\
                         & Config. 2: Reduce and follow\\
                         & Config. 3: Reduce, follow, and overtake\\
    \end{tabular}
\end{table}

For the sake of space, the description of experiments is provided only for the most challenging configurations, namely VRU-1 config. 3, VRU-2 config. 2, and VRU-3 config. 3. Data showing a summary of results obtained for all the use cases in their different configurations are provided at the end of the section. 

\subsubsection{VRU-1 Configuration 3: pedestrian walks in parallel, turns head towards the car, and crosses the street}
\begin{figure}[ht]
\centering
    \includegraphics[width=1.0\linewidth]{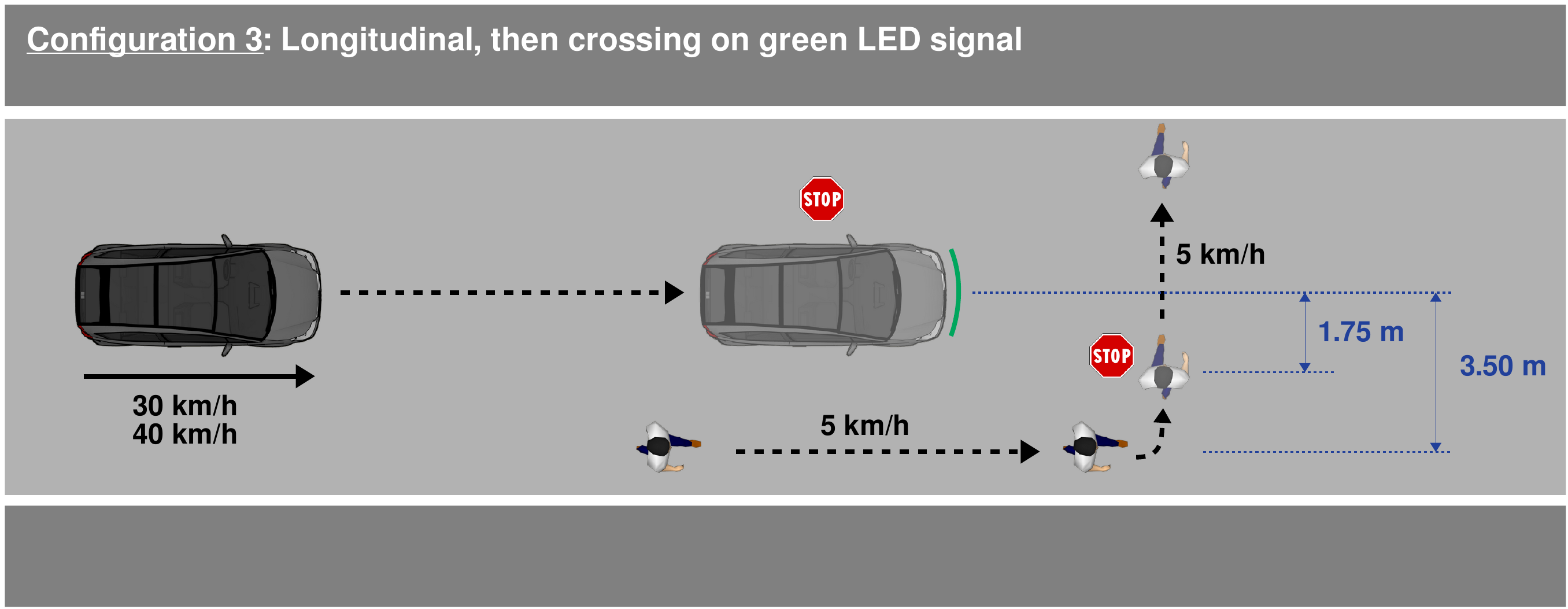}
    \caption{Use case VRU-1 Conf. 3 - Walk, stop, and cross.}
    \label{fig:vru-1-3}
\end{figure}

These tests were conducted using a dummy with an articulated head that can be turned to emulate that the pedestrian is looking at the driver before starting to cross the street. Initially, the pedestrian walks in parallel with the road. After a while, the pedestrian stops and turns his head towards the car, emulating eye contact with the driver. The \ac{VUT} is then expected to detect the pedestrian’s intention to cross the street and, consequently, to decrease speed, gradually and smoothly, until coming to a full stop. At the same time, the \ac{VUT} automatically activates the GRAIL system on, as a means to signal to the pedestrian that the \ac{VUT} has detected their intention to cross, and it is going to brake. The graphical representation of this configuration is depicted in Fig. \ref{fig:vru-1-3}.

\begin{figure}[ht]
\centering
    \includegraphics[width=1.0\linewidth]{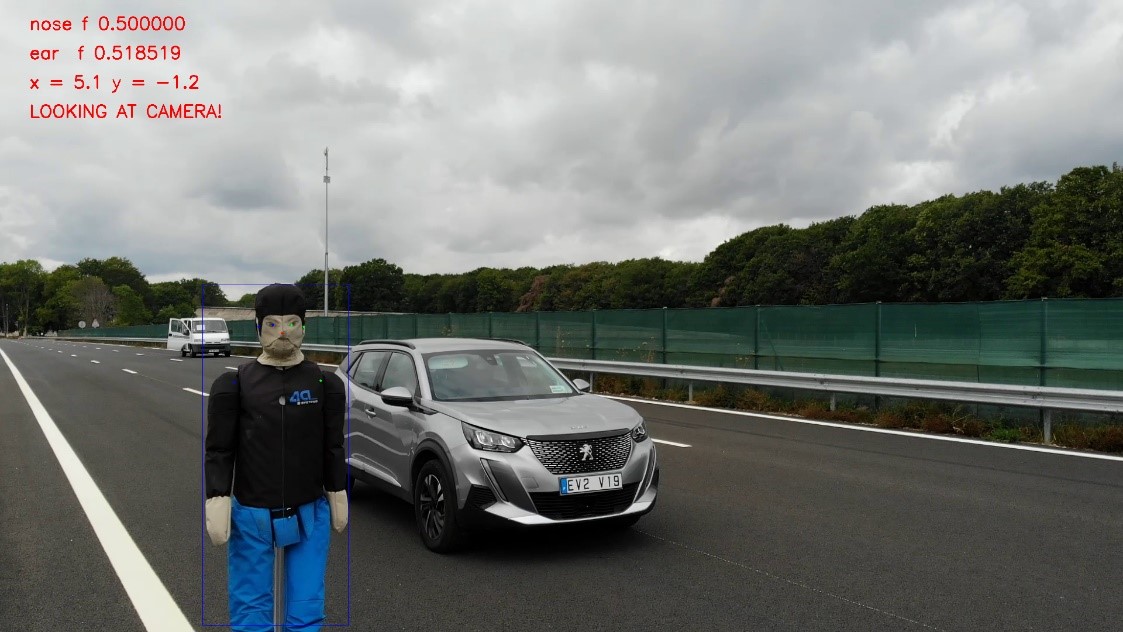}
    \caption{Detection of dummy face using computer vision.}
    \label{fig:detection_dummy}
\end{figure}

Fig. \ref{fig:detection_dummy} depicts an example of how the computer vision system detects the dummy face. The green dots overlaid on the face represent the face features recognized by the algorithm. These features are used to find out whether the dummy is looking at the \ac{VUT}. 

\begin{figure}[ht]
\centering
    \includegraphics[width=1.0\linewidth]{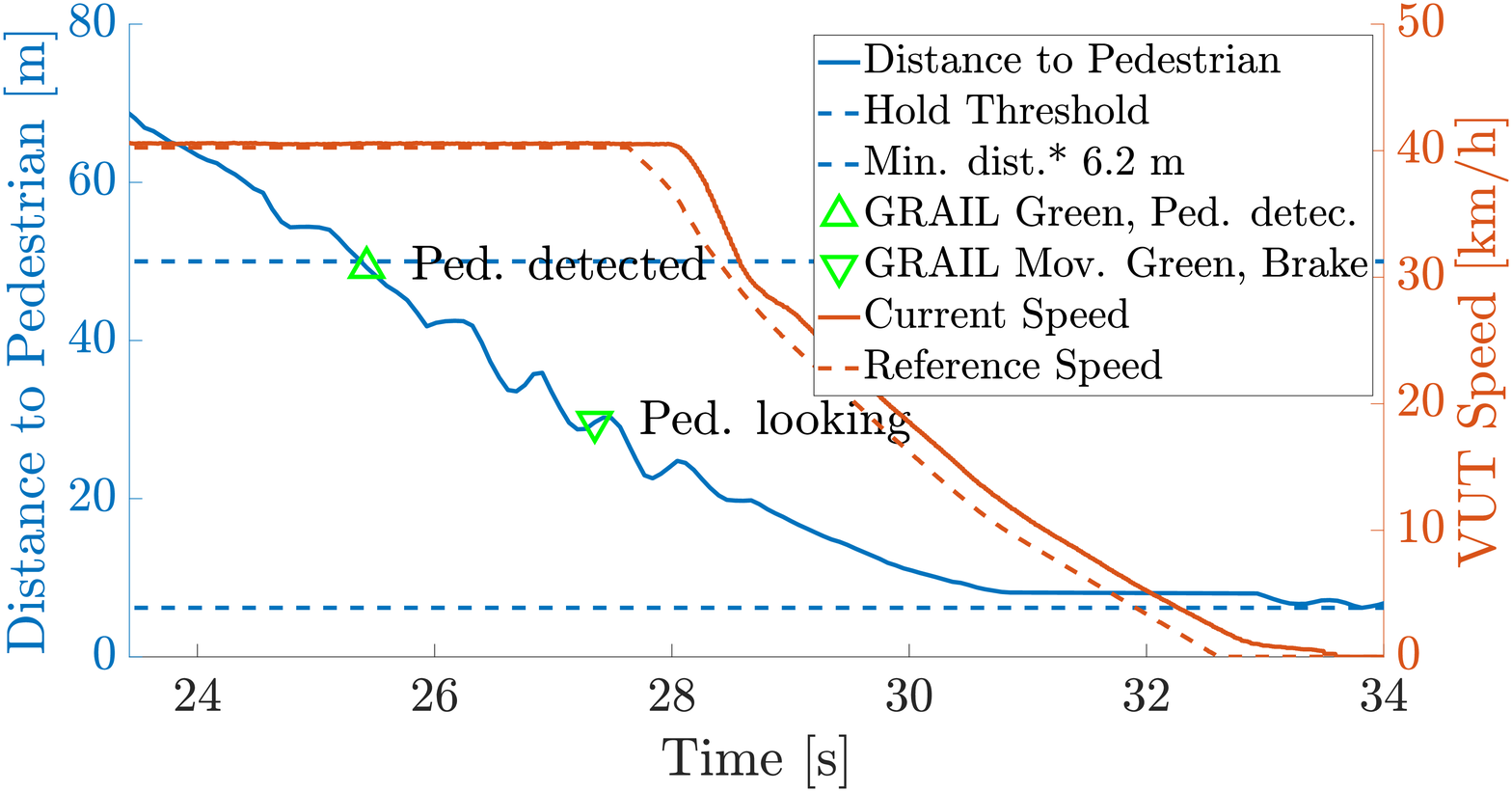}
    \caption{Data logged in VRU-1 Conf. 3 at 40 km/h.}
    \label{fig:datalog_vru1_conf_3}
\end{figure}

Fig. \ref{fig:datalog_vru1_conf_3} shows the data logged during the experiments conducted at 40 km/h. In the figure, the longitudinal distance between the \ac{VUT} bumper and the pedestrian (B2P – Bumper to Pedestrian) and the \ac{VUT} speed are depicted in blue and orange, respectively. 

The dashed orange curve shows the \ac{VUT} reference speed, while the solid orange curves depict the \ac{VUT} real speed. The figures also show the moments of pedestrian detection (pedestrian considered) and face detection (pedestrian looking) by means of green triangles. The first triangle determines the moment when the GRAIL system is preactivated (even though the pedestrian is first detected well in advance by the onboard camera). The second triangle determines the moment when the pedestrian face is looking at the driver. It can be clearly observed that the \ac{VUT} speed starts to diminish as soon as the GRAIL system is switched on. The \ac{VUT} comes to a full stop in front of the pedestrian at a distance ranging between 6-7 meters, providing safety and a comfortable margin for the pedestrian to cross. The quick decisions taken by the system lead to anticipative maneuvers that allow to perform smooth and safe actions. Once more, this way of operation is expected to contribute to increasing the feeling of comfort in the \ac{VUT} passengers, and the feeling of safety and respect in pedestrians.

\subsubsection{VRU-2 Configuration 2: occluded pedestrian crossing at 50\% impact (stop)}
\begin{figure}[ht]
    \centering
    \begin{subfigure}[b]{0.48\linewidth}
        \centering
        \includegraphics[width=\linewidth]{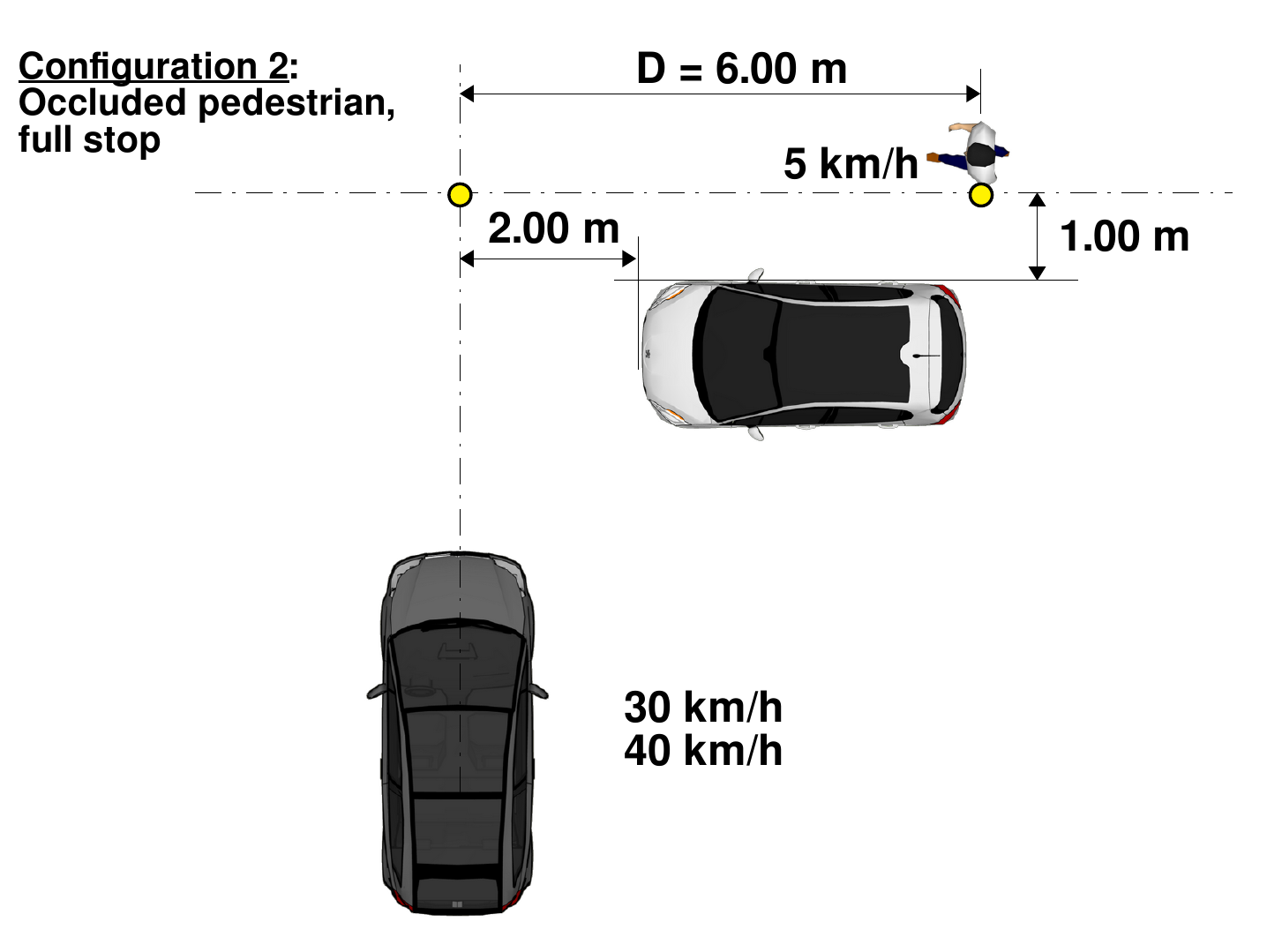}
        \caption{Graphical representation.}
        \label{fig:vru2-conf2-description}
    \end{subfigure}
    \hfill
    \begin{subfigure}[b]{0.48\linewidth}
        \centering
        \includegraphics[width=\linewidth]{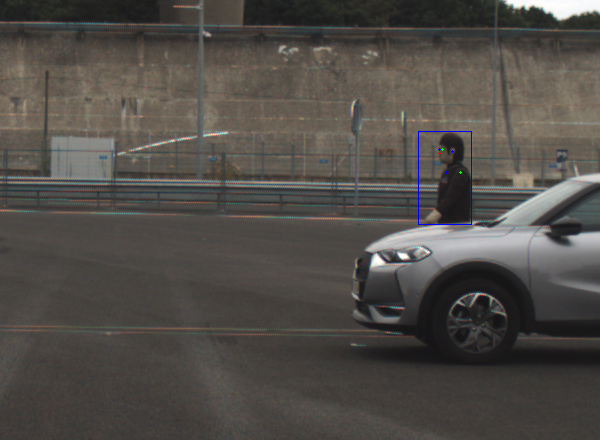}
        \caption{VRU detection example.}
        \label{fig:vru2-conf2-example}
    \end{subfigure}
    \caption{Use case VRU-2 Configuration 2 – Occluded pedestrian crossing at 50\% impact (stop).}
    \label{fig:vru2-conf2}
\end{figure}

This test was conducted with a dummy emulating a pedestrian that emerges from behind a parked car and starts to cross the street unexpectedly. As in the previous configuration, the trajectory of the pedestrian is estimated to intersect the VUT trajectory (if both the VUT and the pedestrian keep a constant velocity) at the central point of the VUT. Thus, the overlap between the VUT and the pedestrian at the time of impact is estimated to be 50\%. This test was conducted at two different VUT speeds: 30 and 40 km/h. The pedestrian walks at a constant velocity of 5 km/h with an initial lateral offset of 6 m with respect to the main axis of the VUT. At the beginning of the test, the pedestrian is not visible from the VUT, given that it is occluded by a parked car. Fig. \ref{fig:vru2-conf2-description} shows a graphical representation of this configuration, while Fig. \ref{fig:vru2-conf2-example} shows a snapshot of the dummy being detected by the computer vision algorithm during the execution of the test. As can be observed, the detected pedestrian (highlighted by a blue bounding box) is only partially visible from the VUT given that it is occluded by the parked car. 

\begin{figure}[ht]
    \centering
    \begin{subfigure}[b]{\linewidth}
        \centering
        \includegraphics[width=1.0\linewidth]{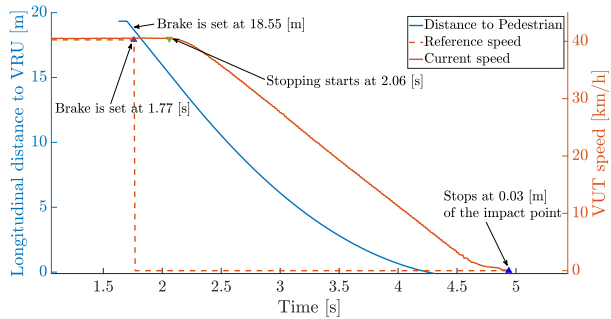}
        \caption{VUT speed and VRU distance.}
        \label{fig:vru2-conf2-speed}
    \end{subfigure}
    \hfill
    \begin{subfigure}[b]{\linewidth}
        \centering
        \includegraphics[width=1.0\linewidth]{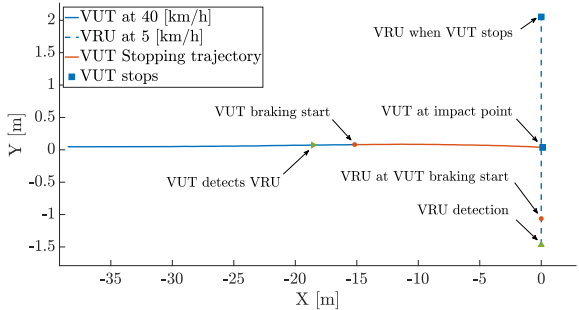}
        \caption{VUT and VRU trajectories.}
        \label{fig:vru2-conf2-trajs}
    \end{subfigure}
    \caption{VRU-2 Configuration 2 – Results at VUT = 40km/h.}
    \label{fig:vru2-conf2-results}
\end{figure}

The early detection of the upper body of the pedestrian makes it possible to start the reaction maneuver in time. Fig. \ref{fig:vru2-conf2-results} provides a graphical representation of the main variables measured during the execution of the test with  VUT speed of 40 km/h. Fig. \ref{fig:vru2-conf2-speed} shows the VUT-VRU longitudinal distance (on the left) in blue, and the VUT speed (on the right) in orange. Both the reference VUT speed (solid) and the current speed (dashed) are provided. As observed, the VUT reference speed is set to zero as soon as the pedestrian is detected. This is intended to execute the braking maneuver as abruptly as possible, given that the situation is critical, and safety is the only variable to consider. The VUT is capable of coming to a full stop at a distance of just a few centimeters (3 cm) in front of the pedestrian. The VUT cannot start the braking maneuver until the pedestrian starts to be perceived by the onboard sensors. This happens with certain delay given the occlusion conditions (see details in fig \ref{fig:vru2-conf2-trajs}). In these circumstances, it can be stated that a VUT velocity of 40 km/h is the limit to guarantee safety in this typical urban situation. As a consequence, the recommendation to set a speed limit of 30 km/h in urban areas is fully supported by the results obtained in this test.

\subsubsection{VRU-3 Configuration 3 – Reduce, follow, and overtake}
\begin{figure}[ht]
\centering
   \includegraphics[width=1.0\linewidth]{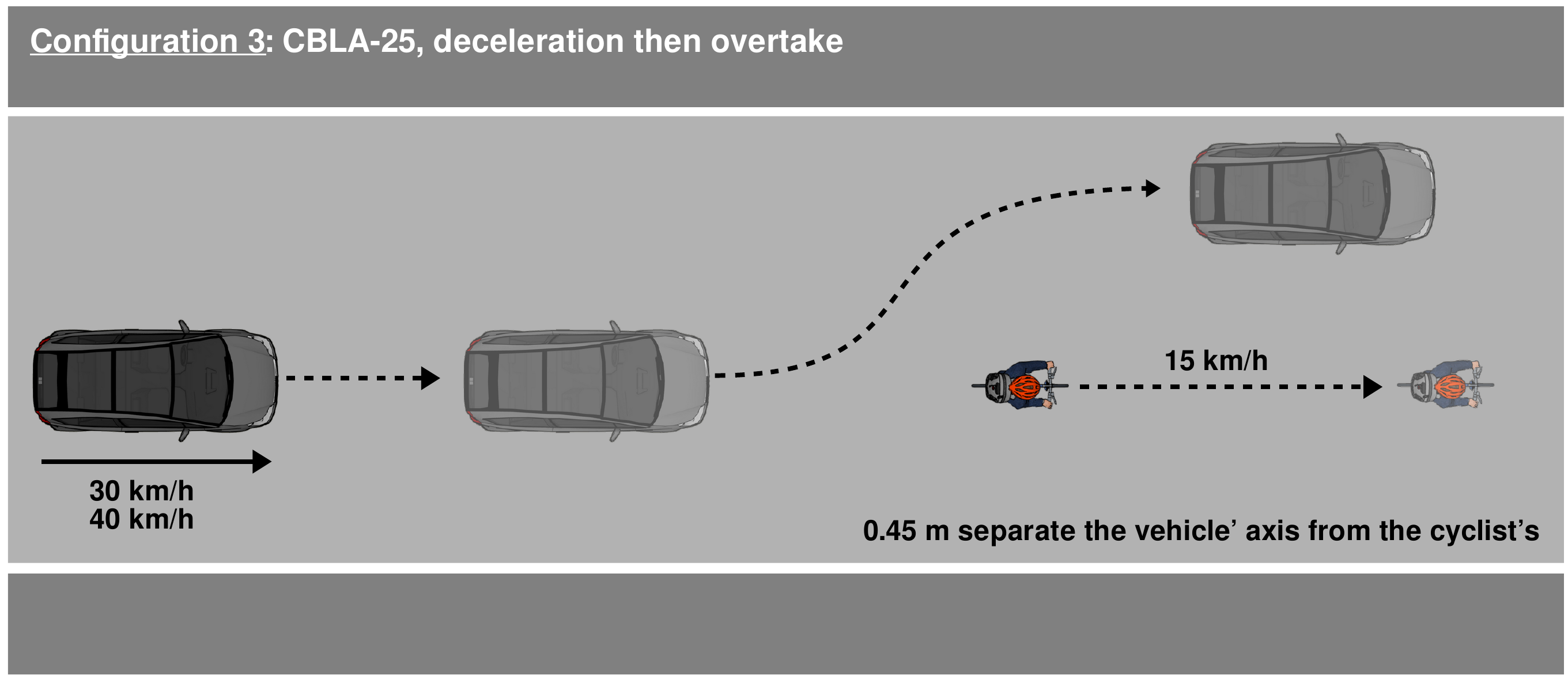}    
    \caption{Use case VRU-3 Conf. 3-Reduce, follow, and overtake.}
    \label{fig:vru3-conf3}
\end{figure}

\begin{figure}[ht]
\centering
    \includegraphics[width=1.0\linewidth]{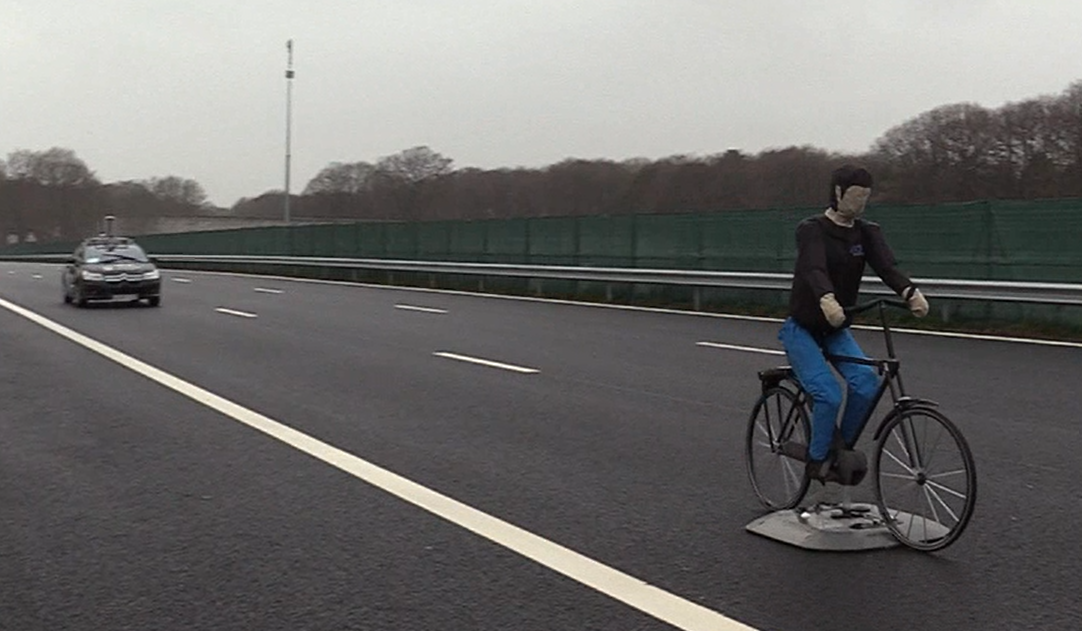}
    \caption{Cyclist dummy used in VRU-3 experiments.}
    \label{fig:vru3-dummy}
\end{figure}

In this configuration, the VUT approaches the cyclist, observes that there is oncoming traffic on the adjoining lane, and waits until the lane is free. After that, the VUT starts to execute a smooth overtaking maneuver with sufficient lateral distance with respect to the cyclist to ensure comfort and safety simultaneously. The graphical description of this configuration is depicted in Fig. \ref{fig:vru3-conf3}. The VUT moves at a velocity ranging 30/40 km/h. The cyclist moves at a constant speed of 15 km/h. The initial lateral offset between the VUT main axis and the cyclist trajectory is 0.45 m. Fig. \ref{fig:vru3-dummy} shows the cyclist dummy used in the experiment. 

\begin{figure}[ht]
\centering
    \includegraphics[width=1.0\linewidth]{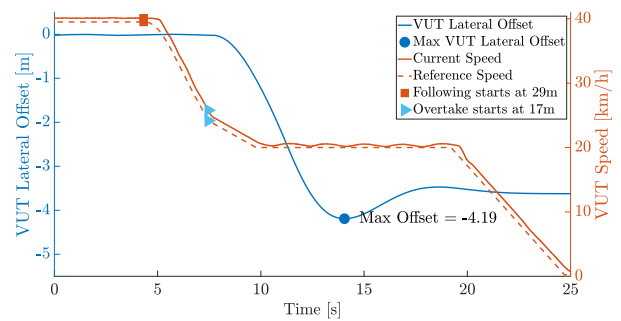}
    \caption{VRU-3 Conf. 3 – VUT speed and lateral offset.}
    \label{fig:vru3-conf3-results}
\end{figure}

Fig. \ref{fig:vru3-conf3-results} shows the VUT reference (blue dashed) and current (blue solid) speed and the VUT lateral offset (in orange), respectively. As can be observed, the VUT performs a smooth maneuver divided in three steps. In the first step, the VUT keeps a constant speed of 40 km/h. In a second step, the VUT decreases velocity until reaching a value of 20 km/h that is kept constant for a while until the oncoming traffic disappears. In the final step, the VUT performs the overtaking maneuver while keeping a constant speed of 20 km/h and leaving a lateral safety distance of 4 m with respect to the cyclist. As in the previous configurations, the VUT is not allowed to initiate the overtaking maneuver until the distance with the cyclist reaches a predetermined value and no predictive system was implemented, given that it was considered to provide no added value in the conditions specified in the test. The passengers’ comfort and the cyclist’s safety are once more guaranteed due to the early detection and smooth action. The conclusion of these experiments with cyclists is that, although comfort and safety are guaranteed, the AI-based predictive systems can offer much more for the sake of safety and anticipation in critical situations. However, it would be necessary to further develop the dummy technology to enable additional dummy movements (turning head, raising arm, inclining the body to the left or to the right, changing the pedaling pace, etc.). The deployment of such technology would allow the testing of much more challenging cyclist-based use cases.   

\subsection{Assessment of Vehicle-Vehicle Interaction}\label{subsec:veh}
Several tests were conducted in order to assess Vehicle-Vehicle interactions following different configurations, as described in table \ref{table:description-veh-1-veh-2}. The schematic description of all the use cases tested is provided as follows.
\begin{itemize}
    \item VEH-1 Cut-in: The VUT interacts with other vehicles willing to merge the mainstream lane from an entry ramp. 
    \item VEH-2 Intersections: The VUT interacts with other vehicles entering an intersection.
\end{itemize}

\begin{table}[ht]
\renewcommand{\arraystretch}{1.2}
    \centering
    \caption{Description of configurations in use cases VEH-1 and VEH-2.}
    \label{table:description-veh-1-veh-2}
    \begin{tabular}{l|l}
    \textbf{VEH Use Case}           & \textbf{Configuration} \\
    \hline
    VEH-1                           & Configuration 1: reduce and follow\\
                                    & Configuration 2: overtake\\
    \hline
    VEH-2                           & Configuration 1: crossing at intersection\\
                                    & Configuration 2: turning at intersection\\
    \end{tabular}
\end{table}

As in the previous section, the description of experiments is provided only for the most challenging configurations, namely VEH-1 config. 2, and VEH-2 config. 2. Data showing a summary of results obtained for all the use cases in their different configurations are provided at the end of the section.

\subsubsection{VEH-1 Configuration 2 – Change lane}
\begin{figure}[ht]
    \centering
    \includegraphics[width=1.0\linewidth]{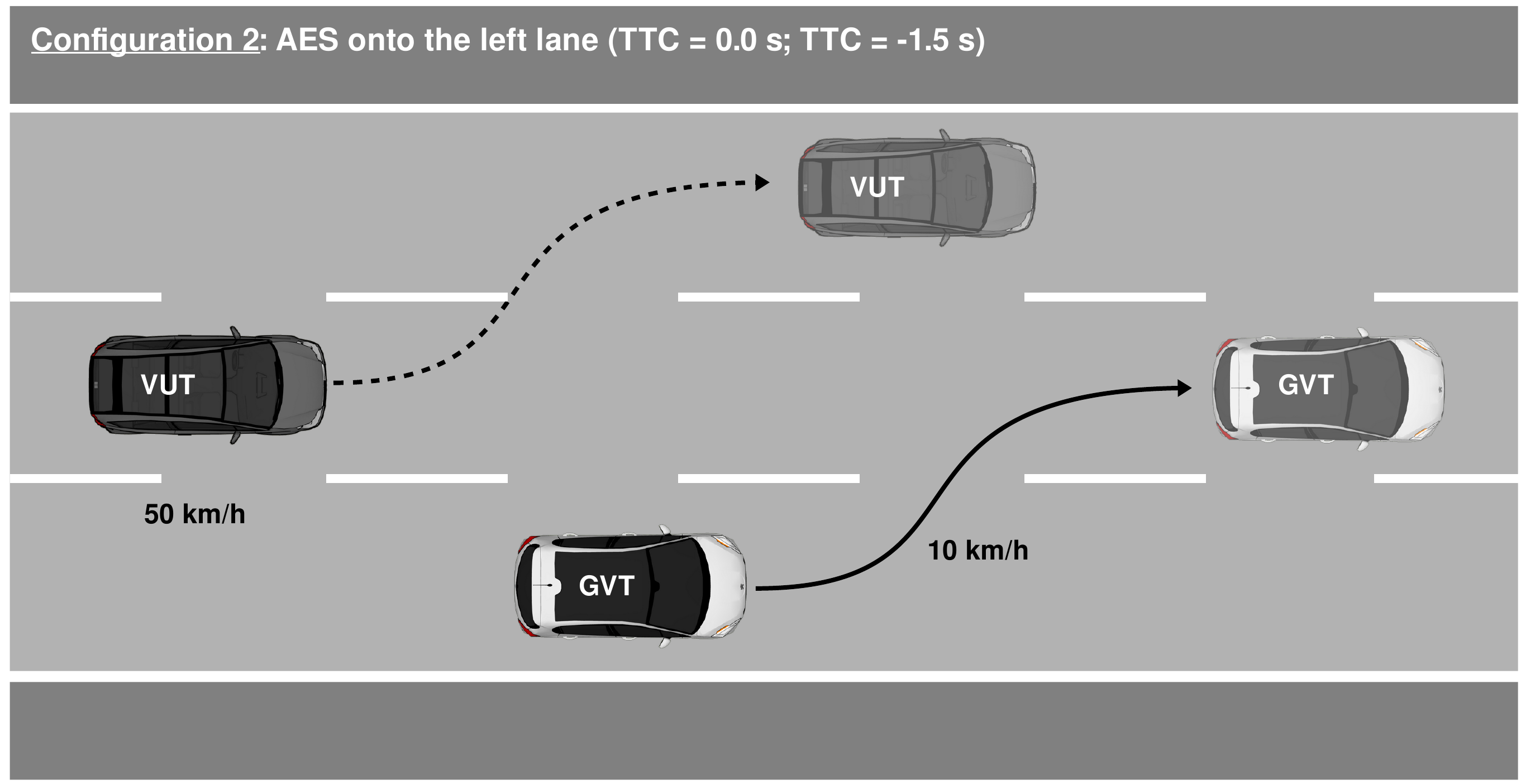}   
    \caption{Use case VEH-1 Configuration 2 – Overtake.}
    \label{fig:veh1-conf2}
\end{figure}

\begin{figure}[ht]
    \centering
    \includegraphics[width=1.0\linewidth]{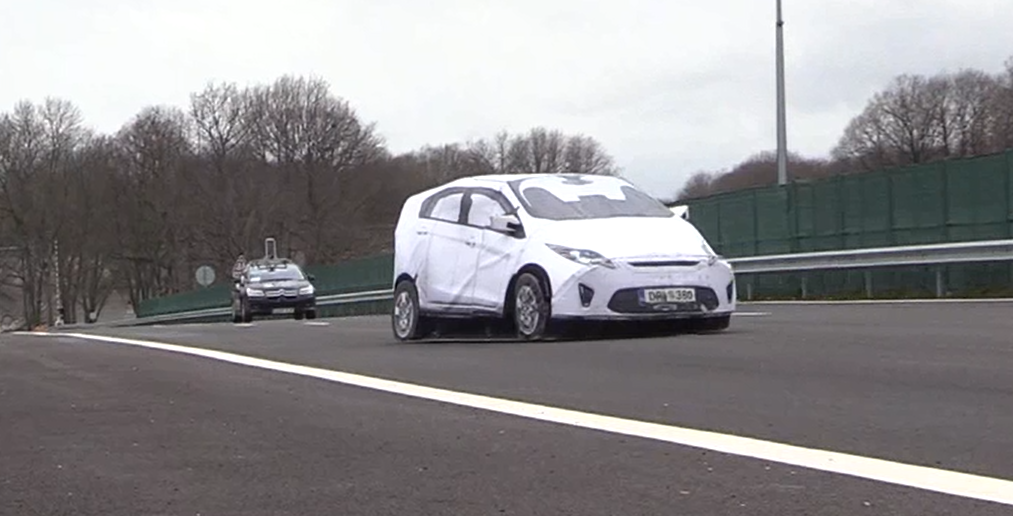}
    \caption{Vehicle dummy used in VEH use cases.}
    \label{fig:veh-dummy}
\end{figure}

Two different configurations were tested in this use case. Both configurations simulate a vehicle entering a highway from an entry ramp. In both cases, the trajectories of the \ac{VUT} and the merging vehicle must be synchronized in order to generate a smooth maneuver. The \ac{VUT} drives at a constant speed of 50 km/h. The \ac{GVT} drives at a constant speed of 10 km/h. These relative speeds simulate a common highway scenario in which the main traffic flow drives at 120 km/h and the side traffic flow merges at 80 km/h. The first configuration (Configuration 1) evaluates the Adaptive Cruise Control (ACC) functionality assuming that the \ac{GVT} will merge in front of the \ac{VUT} and there is no chance to use the adjoining lane. The \ac{VUT} should start to brake as soon as it considers that the \ac{GVT} will merge into the \ac{VUT}’s trajectory. The second configuration (Configuration 2) evaluates the Automatic Emergency Steering (AES) functionality assuming that the \ac{GVT} will merge in front of the \ac{VUT}, and the adjoining lane is available to be used in a lane change maneuver. The \ac{VUT} should start to change the lane as soon as it considers that the \ac{GVT} will merge into the \ac{VUT}’s trajectory. The \ac{VUT}'s performance relies on the prediction or detection of an oncoming lane change in both configurations. The earlier the lane change is detected the longer the time gap and the relative distance to the \ac{GVT} will be. This anticipation will allow the implementation of smooth actions, leading to higher safety and comfort levels. 
The graphical description of this configuration is depicted in Fig. \ref{fig:veh1-conf2}, while Fig. \ref{fig:veh-dummy} shows the vehicle dummy used in the experiments.

In addition, two different use cases were applied for each configuration, triggering the merging maneuver by the \ac{GVT} as a function of the Time-to-Collision (TTC) between the \ac{VUT} and the \ac{GVT} once the merging maneuver by the \ac{GVT} is completed (the time that the \ac{GVT} needs to perform the lane change maneuver is fixed and known). The TTC means the remaining time before the \ac{VUT} would strike the \ac{GVT} assuming that both the \ac{VUT} and \ac{GVT} would continue to travel with the speed it is traveling. For the first case, a TTC$=0$ seconds was used. That is, the GVT initiates the merging maneuver to supposedly end it with its rear bumper in contact with the front bumper of the \ac{VUT}, if none of the vehicles modify their speed and assuming in this case a relative speed of 40 km/h. The second case was defined to be more challenging, with a TTC$=-1.5$ seconds (in this case negative TTC values correspond to a post-collision situation). That is, the \ac{GVT} starts the merging maneuver so as to supposedly end it with a relative distance from the front bumper of the \ac{VUT} to the rear bumper of the \ac{GVT} of -16.67 meters for a relative speed of 40 km/h. In other words, the time that the \ac{VUT} would have to react in this second case is 1.5 seconds less than in the first case. Both cases would result in a collision if the \ac{VUT} took no action. By using the TTC between the \ac{VUT} and the \ac{GVT} at the end of the merging maneuver of the \ac{GVT} these scenarios can be easily adapted to different distances and speeds.


As already mentioned, the relative velocity between the \ac{VUT} and the \ac{GVT} is 40 km/h, emulating two vehicles driving on a highway at 120 km/h (\ac{VUT}) and 80 km/h (\ac{GVT}), respectively. On this occasion, the \ac{VUT} checks the adjoining lane and determines that there is no oncoming traffic. After that, the \ac{VUT} proceeds to execute a lane change overtaking maneuver in order to leave plenty of room for the \ac{GVT} to merge safely. The further anticipation the smoother the maneuver and the larger the safety margin. This maneuver is denoted as Automatic Emergency Steering (AES).

\begin{figure}[ht]
    \centering
    \begin{subfigure}[b]{\linewidth}
        \centering
        \includegraphics[width=1.0\linewidth]{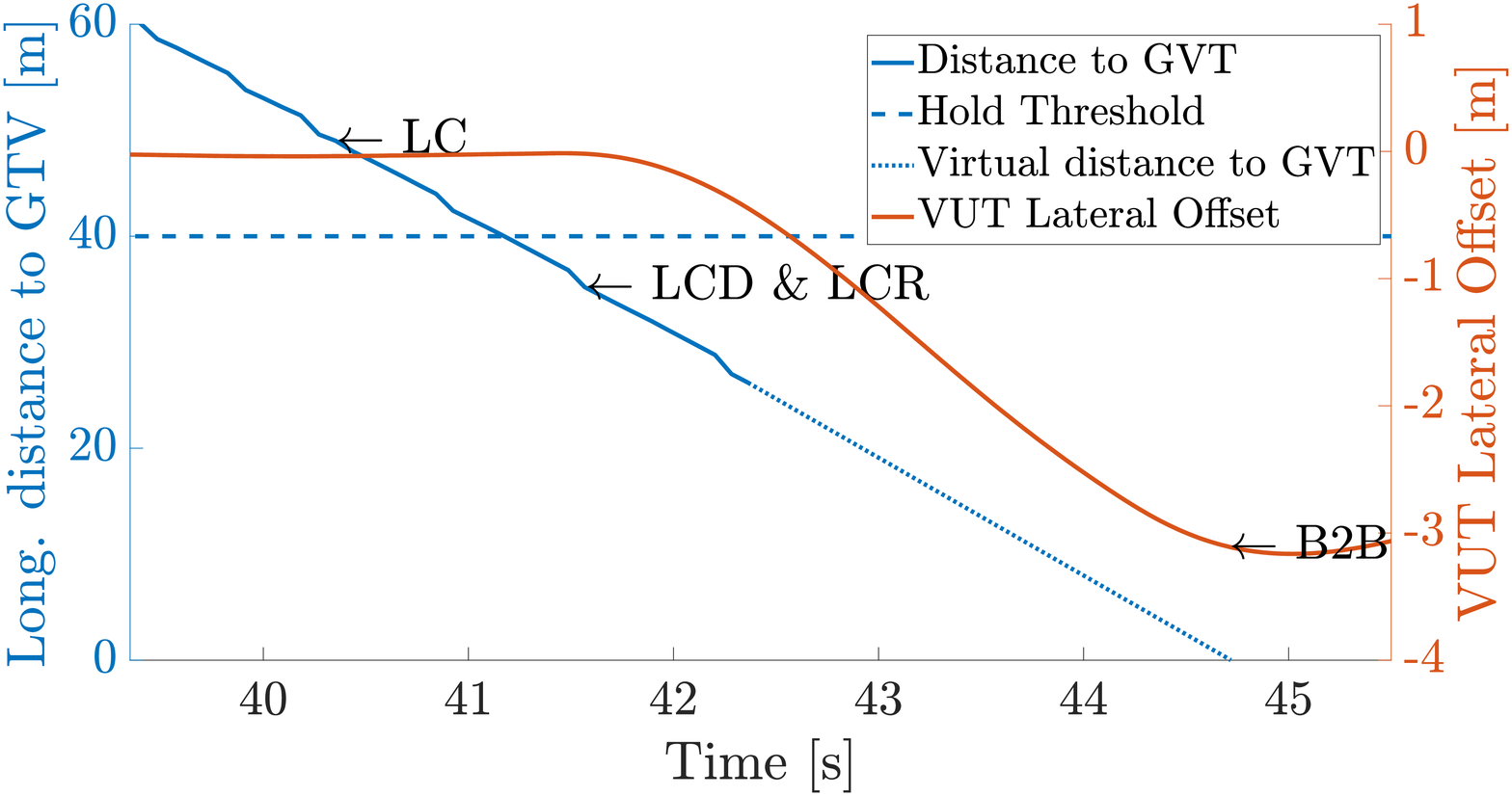}
        \caption{Baseline prediction. LC detection after 1.2 seconds.}
        \label{fig:veh1-conf2-ttc15wopred}
    \end{subfigure}
    \hfill
    \begin{subfigure}[b]{\linewidth}
        \centering
        \includegraphics[width=1.0\linewidth]{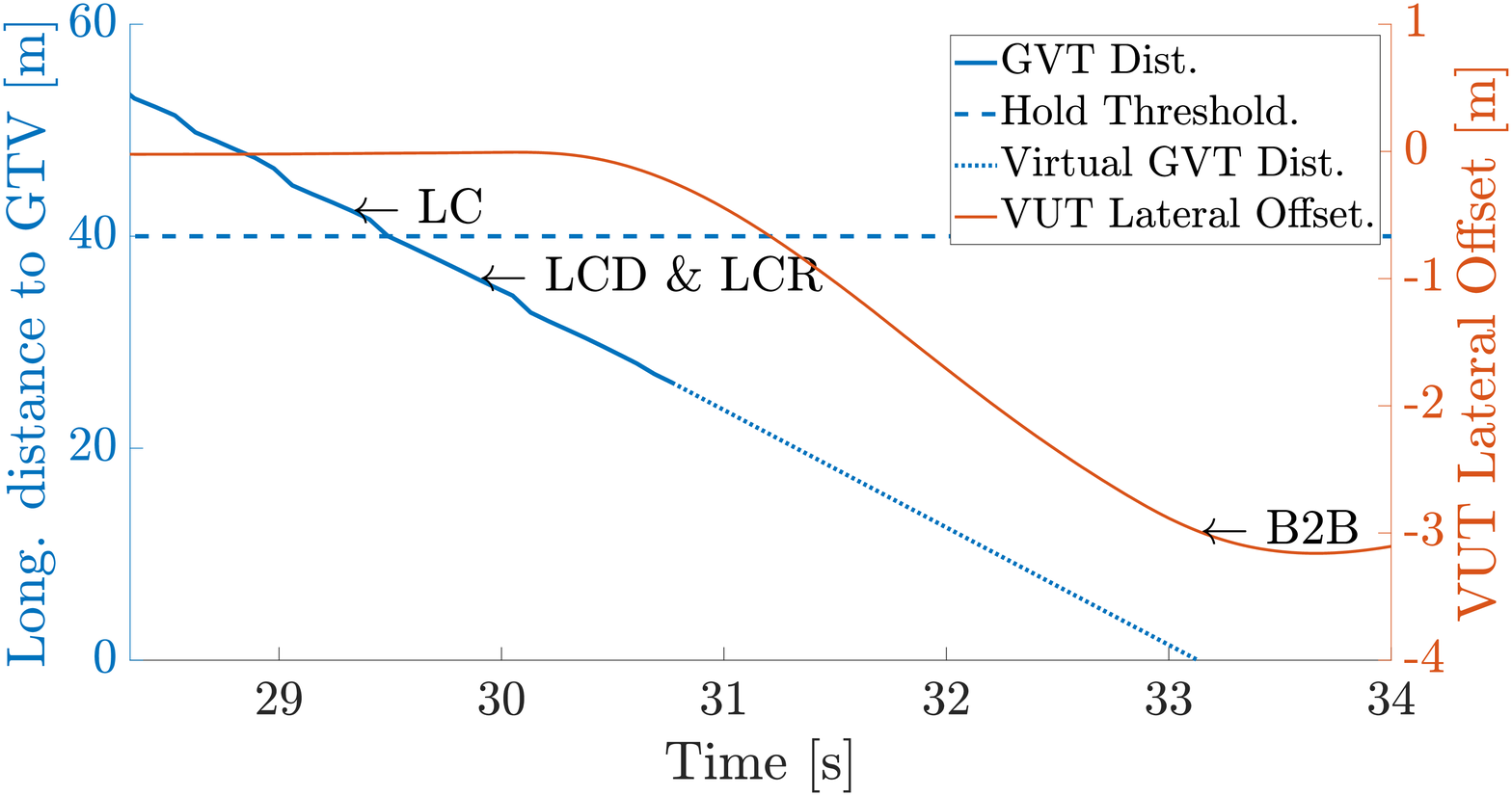}
        \caption{BRAVE prediction. LC detection after 0.5 seconds.}
        \label{fig:veh1-conf2-ttc15wpred}
    \end{subfigure}
    \caption{Use case VEH1 Conf. 2 TTC=-1.5s. Longitudinal distance to GVT and VUT Lateral offset.}
    \label{fig:veh1-conf2_results}
\end{figure}

Fig. \ref{fig:veh1-conf2-ttc15wopred} shows the longitudinal distance between \ac{VUT} and \ac{GVT}, in blue, the estimated distance (dashed) and the lateral offset, in orange, for TTC = -1.5 seconds (the most challenging conditions) using the baseline predictive system based on a Kalman Filter in use case VEH-1 configuration 2. Similarly, Fig. \ref{fig:veh1-conf2-ttc15wpred} shows the same variables using the predictive system developed in BRAVE. Note the different scale in the time axis.

In this case, the reaction time of the baseline system is 1.2 seconds, while the reaction time of BRAVE predictive system is 0.5 seconds. Under these demanding conditions, the difference of anticipation is 0.7 seconds in favor of BRAVE predictive system. This difference is to endow the system with additional reaction time. As a consequence, the predictive system allows to achieve further comfort and safety. The BRAVE predictor leads to higher anticipation times systematically. The average lane change detection time achieved by BRAVE predictor is 0.77 s (very similar to 0.65 s, the result achieved on video sequences after exhaustive testing). This is a bit more than 300ms faster than the average reaction time of humans \cite{iv_2020_challenge}, which is around 1.08 s, and 800 ms faster than the average reaction of the baseline predictive system based on Kalman filtering (1.56 s). Another relevant remark is the fact that BRAVE vehicle scored full points on all use cases similar to EuroNCAP tests.

\subsubsection{VEH-2 Config. 2 – Turning at intersections}
\begin{figure}[ht]
    \centering
    \includegraphics[width=0.9\linewidth]{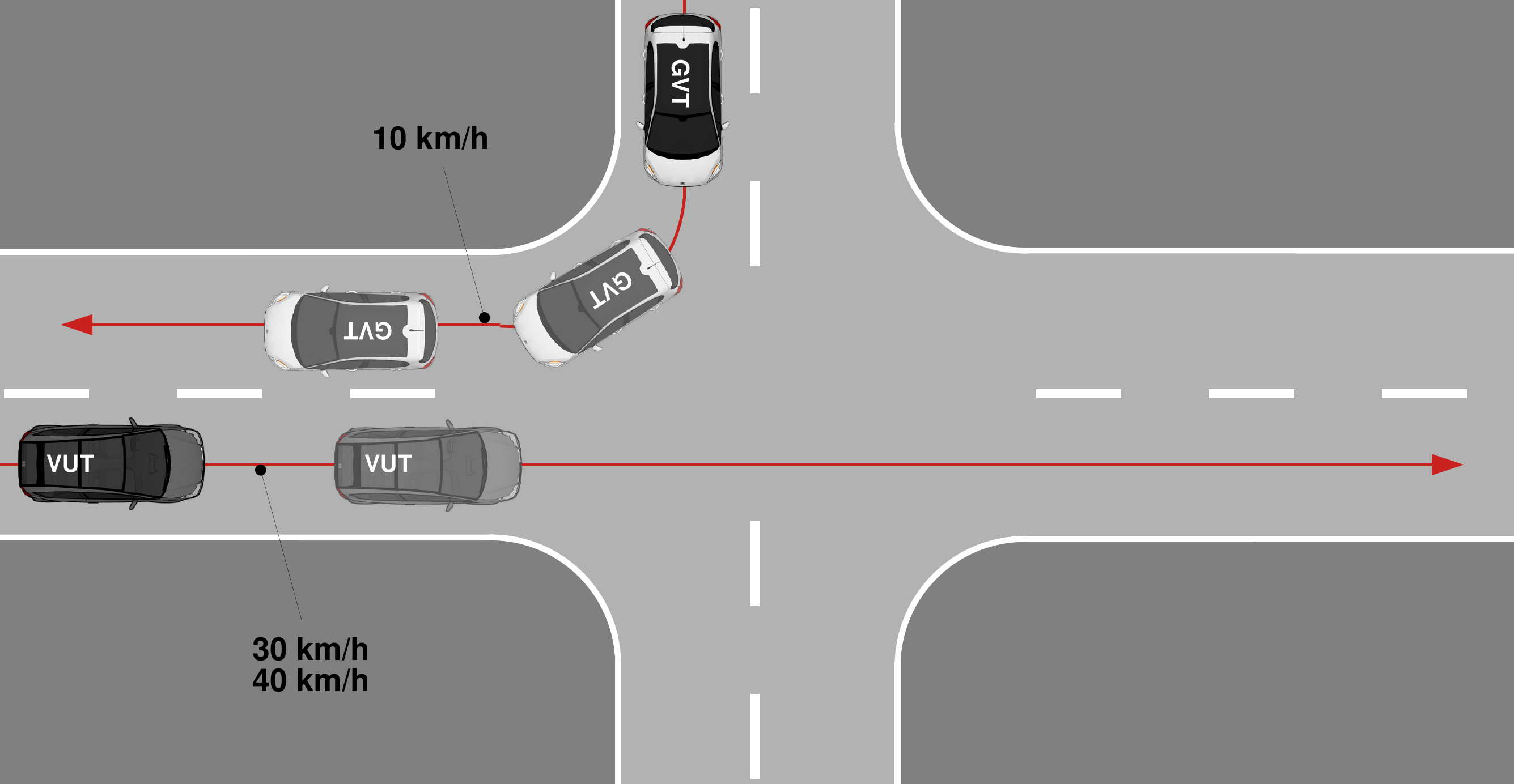}
    \caption{Use case VEH-2 Conf. 2 – Turning at intersection.}
    \label{fig:veh2-conf2}
\end{figure}

Two different configurations were tested in this use case. Both configurations simulate a couple of vehicles, the VUT and the \ac{GVT}, entering an intersection coming from perpendicular directions. In the first case, the \ac{GVT} continues forward at the intersection, cutting the \ac{VUT} trajectory. In such circumstances, the VUT must decelerate and accommodate its speed to give way to the \ac{GVT}. In the second case, the \ac{GVT} turns right at the intersection without interrupting the \ac{VUT} trajectory at all. The \ac{VUT} initially decreases its speed until it predicts the turning maneuver of the \ac{GVT}. In that moment, the \ac{VUT} reference speed is resumed to the cruise value that it had before entering the intersection. This can be seen as a case of false positive detection which, in this case, increases safety in an uncertain situation. Furthermore, the ability to anticipate the turn of the \ac{GVT} allows optimizing the speed of the \ac{VUT} at the intersection. \emph{False positive testing} involves use cases where the GVT finally aborts the expected maneuver. This use case is relevant given that it looks at evaluating systems that can improve the efficiency of traffic at intersections. The graphical description of this scenario is depicted in Fig. \ref{fig:veh2-conf2}.


\begin{figure}[ht]
    \centering
    \begin{subfigure}[b]{\linewidth}
        \centering
        \includegraphics[width=1.0\linewidth]{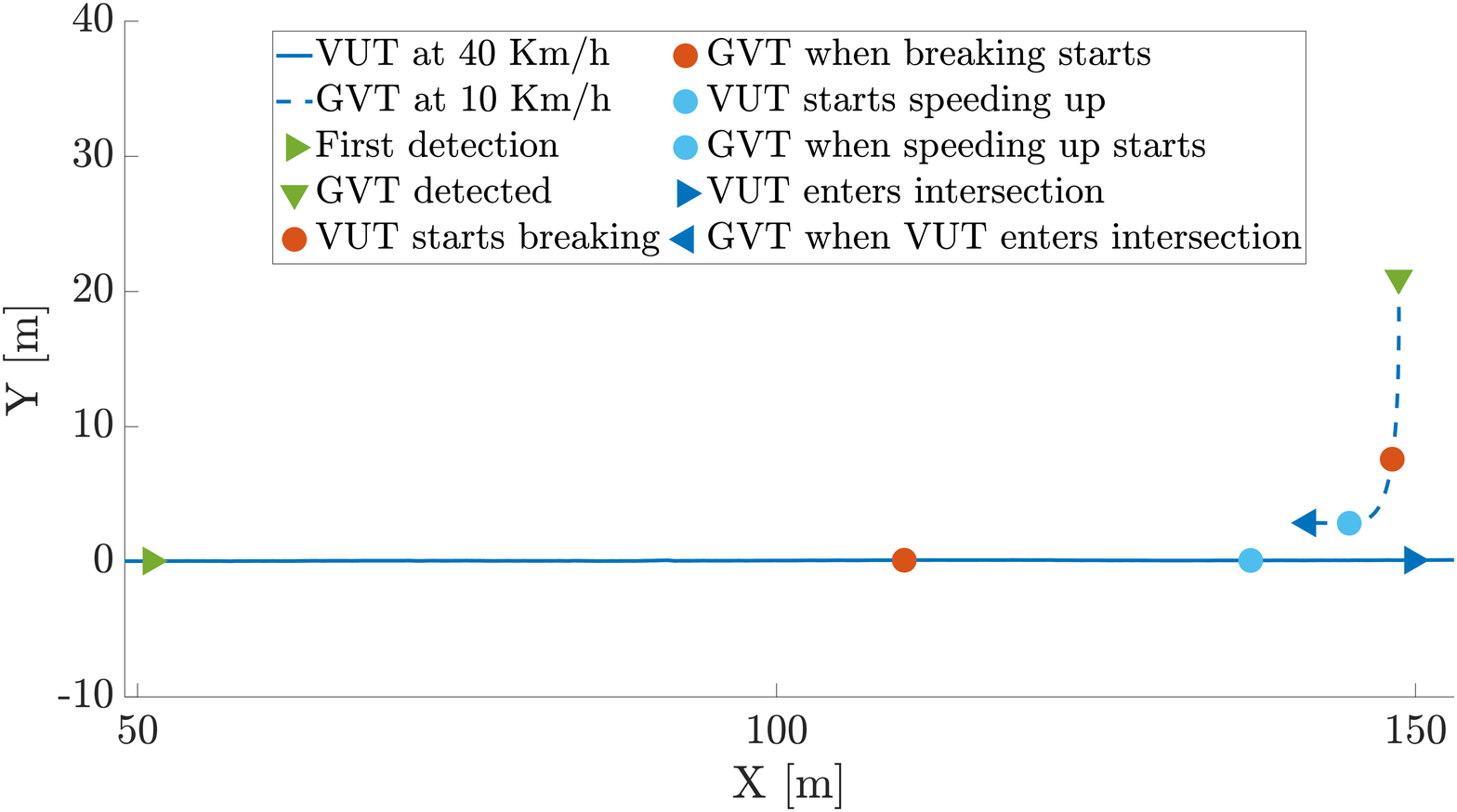}
        \caption{VUT and GVT positions.}
        \label{fig:veh2-conf2-positions}
    \end{subfigure}
    \hfill
    \begin{subfigure}[b]{\linewidth}
        \centering
        \includegraphics[width=\linewidth]{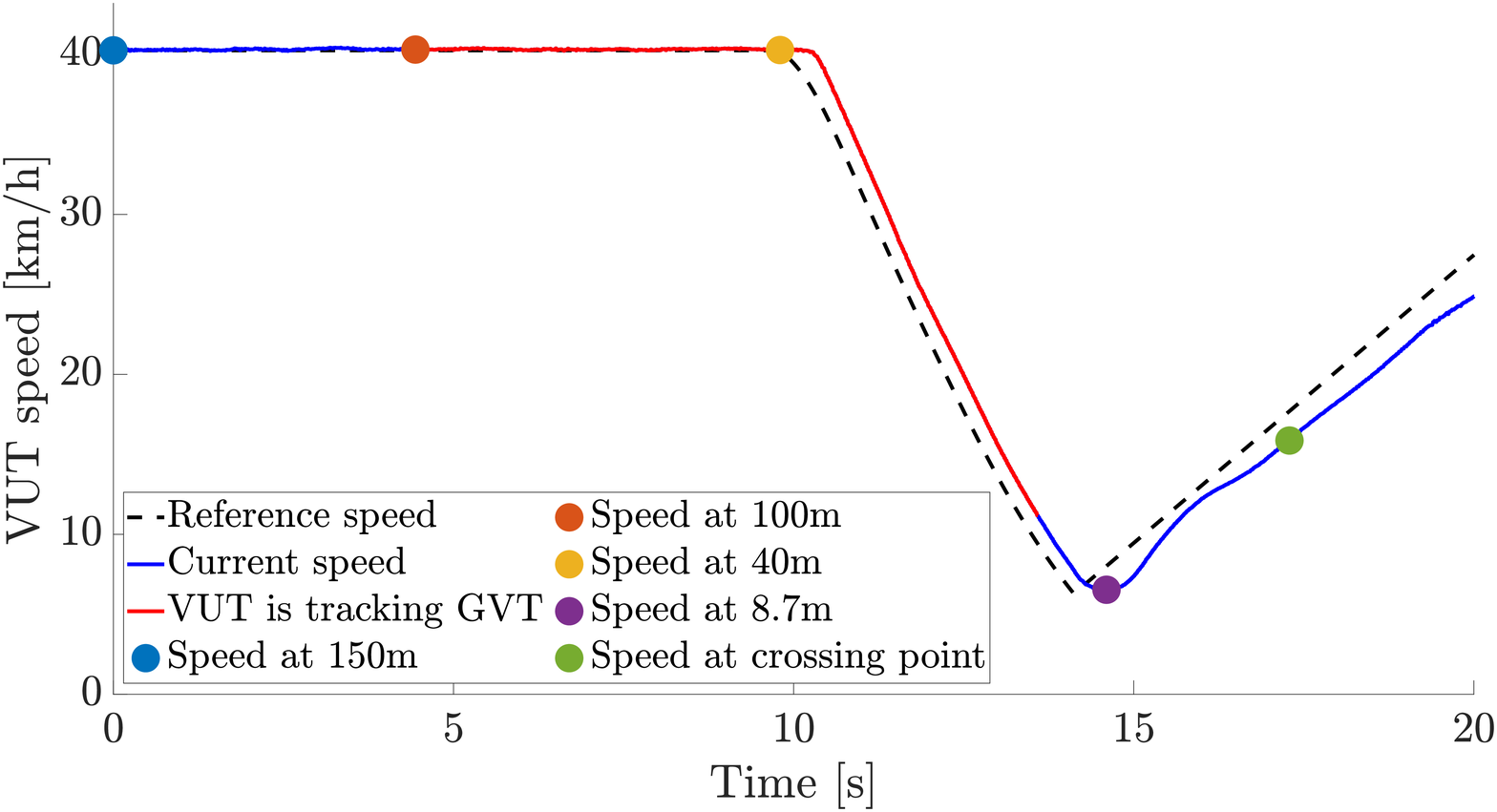}
        \caption{VUT speed (real and reference).}
        \label{fig:veh2-conf2-speed}
    \end{subfigure}
    \caption{Use case VEH-2 Configuration 2.}
    \label{fig:veh2-conf2_data}
\end{figure}

The onboard sensors were used to detect the location of the \ac{GVT} and the relative distance and velocity with respect to the \ac{VUT}. Fig. \ref{fig:veh2-conf2-positions} shows the positions of the \ac{VUT} and \ac{GVT} during the execution of a test with an initial \ac{VUT} speed of 40 km/h. The closest distance between the \ac{VUT} and the \ac{GVT} during the test was again around 4 meters. Fig. \ref{fig:veh2-conf2-speed} shows the \ac{VUT} reference and current speeds. As can be observed, the \ac{VUT} reduces speed following a smooth pattern. The reduction of \ac{VUT} speed is not that big as it was in configuration 1, given that the \ac{VUT} eventually predicts the intention of the \ac{GVT} to turn at the intersection. As soon as the intention is detected (around 2s after starting to decrease speed, marked by an orange square), the \ac{VUT} resumes its reference speed until reaching the cruise speed it had before entering the intersection. Safety and comfort are once more the desired variables to optimize.

This use case has been completely devised in this project, given that it is none of the use cases considered by EuroNCAP. The deployment of this use case allows to issue some recommendations on how to define and execute this type of validation tests in EuroNCAP. The results achieved in these tests, in terms of safety gap and reaction time (time needed to estimate the intention of the \ac{GVT} since it enters the intersection), set a baseline for further testing of more complex intersection use cases in EuroNCAP.
\section{Discussion}
\label{sec:discussion}
%
%
%
%


After addressing the technical description of the tests performed and the results obtained, this section discusses the main findings and the limitations encountered in testing the predictive systems. It also offers potential actions to be implemented in future test procedures. The proposed recommendations are classified according to their difficulty of implementation or time horizon.

\subsection{Main findings}

After the thorough experimentation conducted at UTAC's premises to test the predictive systems developed in the BRAVE project, the following findings can be remarked. 

\begin{itemize}
    \item In general, it can be safely stated that predictive systems provide added value in terms of safety (greater anticipation time, larger safety gap), efficiency, and comfort (smoother maneuvers with minimum jerk and minimal reduction of speed). 
    \item After extensive experimentation in a proving ground under standardized conditions, BRAVE’s predictive systems have proven to outperform the state-of-the-art vehicles tested at UTAC in the same use cases. 
    As a matter of fact, UTAC, as an independent EuroNCAP tester, issued the following qualitative report comparing the performance of BRAVE vehicle and other vehicles with similar capabilities tested on the same use cases \cite{BRAVE_D4.6}:
    \begin{itemize}
        \item \emph{"The BRAVE vehicle tested scored full points on the accomplished tests. EuroNCAP protocols used to obtain scores were written to assess AEB systems."}
        \item \emph{"Comparison between the BRAVE vehicle and other manufacturers’ vehicles with Automated Driving (AD) functions:}
        \begin{itemize}
             \item \emph{BRAVE vehicle is the best performing vehicle as far as relative distance with GVT is concerned for cut-in use cases.}
            \item \emph{BRAVE vehicle was the only vehicle able to avoid any collision fully autonomously for the cut-in -1.5s TTC use case.}
            \item \emph{BRAVE vehicle behaves more smoothly than other vehicles."}
        \end{itemize}
    \end{itemize}

\item BRAVE’s predictive system overcomes humans’ prediction ability on lane changes by 300 ms, as well as the prediction ability of the baseline predictive system (based on Kalman Filtering) by 800 ms. Thus, it demonstrates the added value of AI-based prediction systems as an asset to achieve higher comfort and safety in automated driving, especially in critical scenarios such as lane changing and merging maneuvers on highways. 

\item BRAVE has proved the added value of prediction of intentions in pedestrian use cases by detecting in advance the pedestrian (dummy) face looking for eye contact with the driver. The combined use of anticipated gaze detection and activation of the GRAIL interface leads to smooth behavior, thus contributing to an enhanced feeling of comfort and respect both for pedestrians and passengers of automated cars. 

\item Last-second reactions have been improved (shortened) in order to increase the ability to deal with the most challenging conditions in pedestrian use cases, both in braking and avoidance scenarios. Predictive systems are useful in scenarios with high visibility (pedestrian approaching the curbstone in an open area while being fully visible from the car), but they provide no added value in extreme cases where the pedestrian is occluded (pedestrian crossing the street while suddenly emerging from behind parked cars). In any case, safety was guaranteed, even in those challenging conditions.  

\item The current predictive systems provide no added value in the cyclist use case, given that the cyclist is detected from a far distance and no abrupt cyclist’s maneuver or reaction is carried out. Although safety and comfort were guaranteed during the tests, more technologically advanced cyclist dummies that more accurately mimic real cyclists’ behavior would be needed in order to make the most out of the potential that predictive systems can offer. These are some of the advanced dummy features that would be needed for such purpose: cyclist raising the arm to signal a change of direction; cyclist turning head (as a clear indication of changing direction); cyclist leaning to the left of right; variable pedaling pace. 

\item Intersection use cases involving two vehicles have been deployed successfully, proving the added value of prediction in false positive cases. However, further research would be needed in order to deal with more complex use cases at intersections involving multiple vehicles. \end{itemize}

\subsection{Main limitations}
\begin{figure}[ht]
 \centering
 \includegraphics[width=0.7\linewidth]{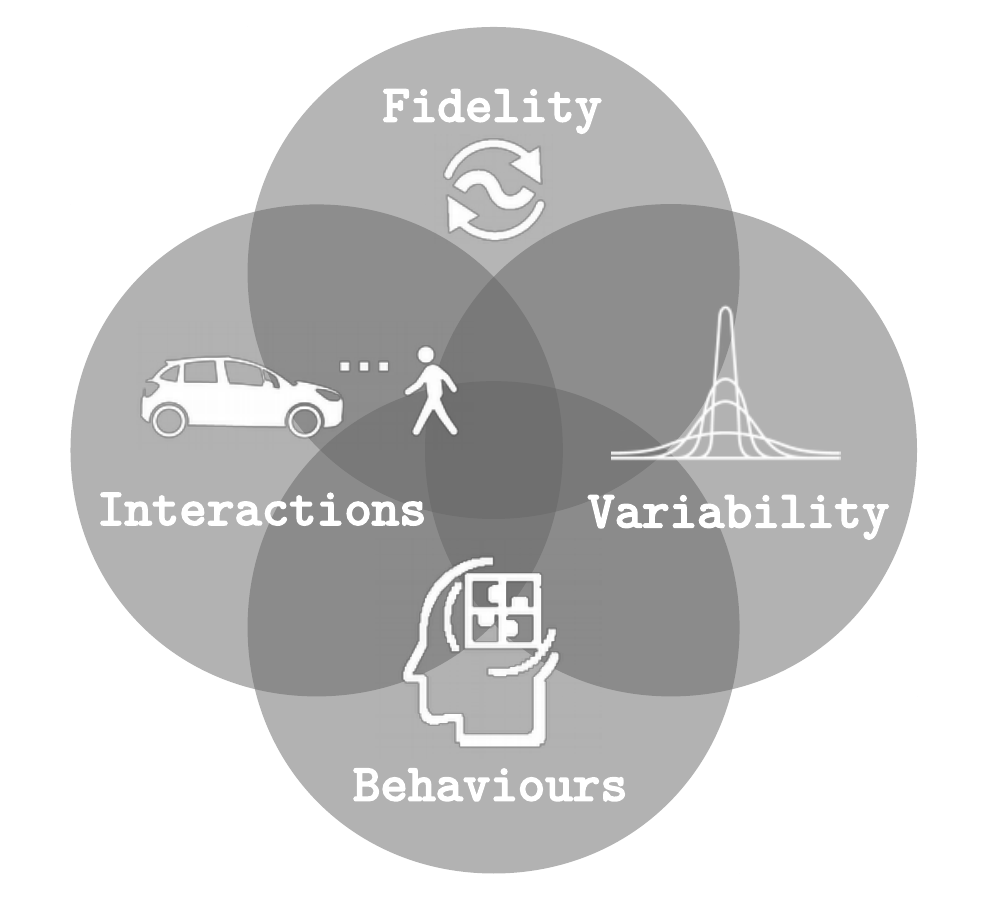}
 \caption{Venn diagram illustrating the main identified weaknesses when testing predictive systems.}
 \label{fig:limitations}
\end{figure}

The main shortcomings identified in the current procedures for testing prediction systems are described below. As illustrated in Fig. \ref{fig:limitations}, it is important to note that these limitations are interrelated and not mutually exclusive.

\subsubsection{Limited fidelity}
\begin{figure}[ht]
 \centering
 \includegraphics[width=0.99\linewidth]{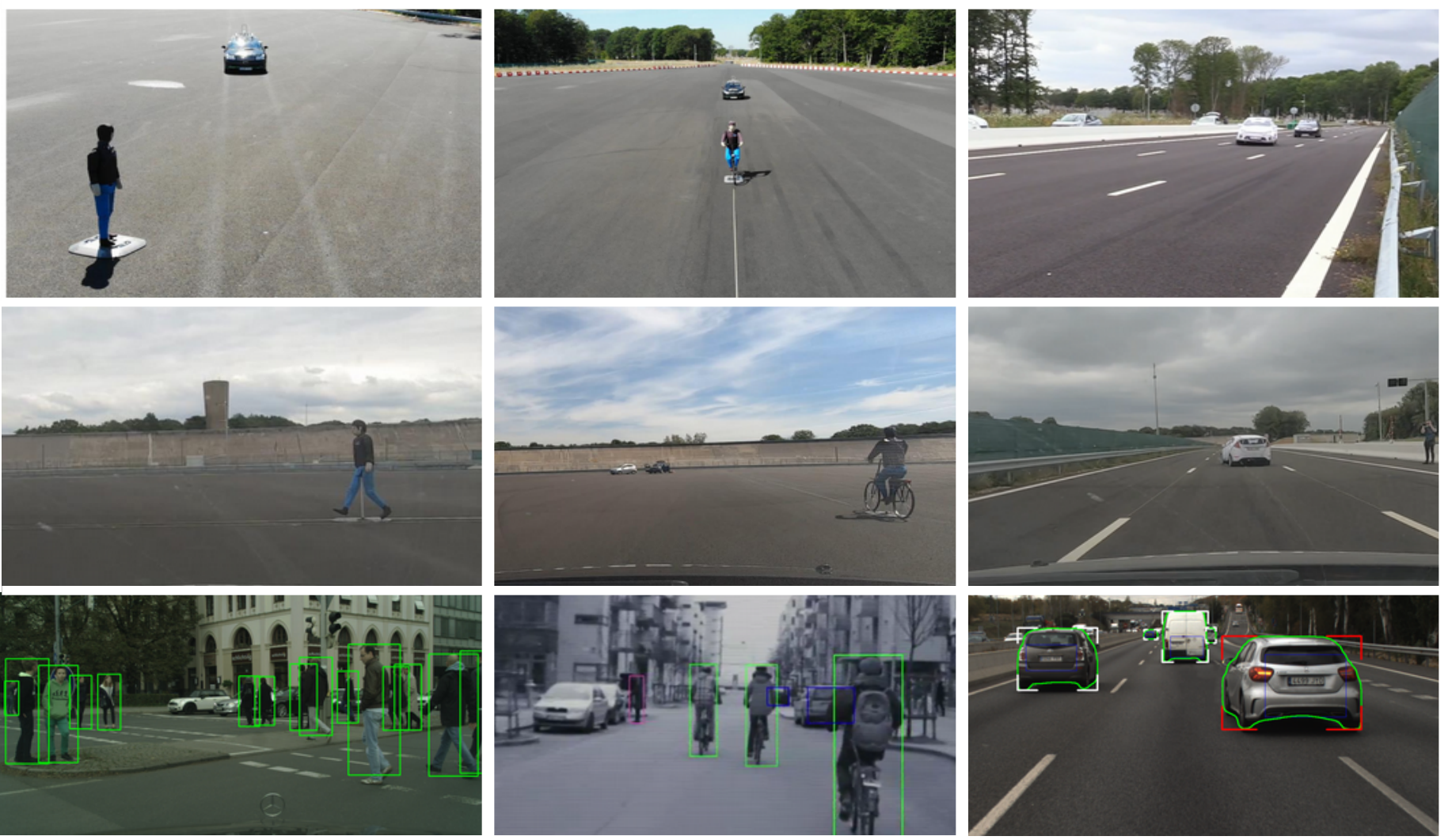}
 \caption{Illustration of the fidelity gap between test and real scenarios. First row: test environment. Second row: view from the vehicle. Third row: examples of real environments.}
 \label{fig:fidelity}
\end{figure}

Standardized tests are easy to carry out, but they do not reflect the behavior of road agents in a realistic manner, since the movements of the robotized dummies are too rigid and linear. Consequently, they are easy to detect and predict. The aspect of the dummies and the GVT is also standardized and thus it is repetitive. Similarly, the testing scenarios are totally open, without any objects in the background, making the test much easier from the perception point of view, whether based on vision, radar, or LiDAR. In addition, only one dummy (human or vehicle) is considered at a time during the tests, making the scenarios too simplistic compared to real driving situations (see Fig. \ref{fig:fidelity}) and eliminating the possibility of including interactions between multiple agents.   



\subsubsection{Limited variability}
The tests are conducted under strict repetitive conditions. Thus, there is little variability in the testing conditions, which are limited to changes of velocity, time-to-collision, and percentage of overlapping area at the point of impact. The potential variability of the appearance of human dummies (e.g., clothing, hair color, skin color, height, etc.), as well as of the dummy vehicle (e.g., color, size, etc.), or even of background conditions (e.g., cluttered background) is not sufficiently exploited. These strict repeatability requirements can lead to tailored solutions that artificial intelligence systems can easily learn. 



%

\subsubsection{Limited behaviors}
The dummies execute standard, pre-programmed behaviors that are limited to very basic actions, such as crossing, not crossing, and changing lane. There are no scenarios representing multi-agent interactions. A larger variation of behaviors, including interactions, is needed for testing advanced predictive systems.  





\subsubsection{Lack of real behaviors}
Most of the systems that have been tested in this project are ”last resort” solutions (i.e., ESP, EBS, ABS, etc.), meaning that the capability to provide mid or long-term anticipation is not tested at all. In addition, the dummies do not perform any reaction as a consequence of the actions of the autonomous vehicle during the tests (e.g., the effect of external HMI systems cannot be assessed). Thus, it is not possible to test the effect of the evaluated systems on other agents’ behavior. Advanced predictive systems should be tested on more realistic circumstances where real interactions between agents take place, as a means to assess the real value that predictive capabilities can bring in this field. 

The above mentioned limitations impose constraints on the validation methods to assess the real potential of predictive systems in the context of automated driving. However, the results obtained suggest that, even with the current certification context, predictive systems provide substantial benefits in terms of anticipation, safety, and comfort. In the following lines, we propose some potential actions intended to pave the way forward in the near future. 

\subsection{Recommendations for future actions}


\subsubsection{Improving fidelity}
Regarding the dummies, we have identified several possible actions. A first proposal is to have robotized dummies performing more realistic trajectories by adding probabilistic noise to the linear trajectories. Another possibility is to modify the baseline trajectories by making them a bit more erratic and, consequently, much less linear and more realistic. A second proposal is to develop dummies that can perform more realistic movements, such as pedestrians bending their upper body forward before starting to walk, pedestrians turning their torso, cyclists turning their head towards oncoming traffic, or vehicles with robotized turn signals. A third proposal is to include multiple dummies in the testing scenarios, so that autonomous vehicles have to reason about several road users and their potential future movements and interactions. 

In terms of environments, any efforts to provide more realistic environments (e.g. realistic urban scenarios such as those used in the DARPA Urban Challenge \cite{DARPA_Urban2009}) would be beneficial to improve fidelity. 






\subsubsection{Improving variability}
The variability of the testing conditions can be largely increased by having the dummies performing different behaviors, apart from the standard crossing, not crossing, etc. Examples of such behaviors are: partially changing trajectory (direction and orientation), stop-and-go, variable velocity, and acceleration profiles, etc. The use of dummies with different conditions of clothing, hair, and skin color can also be interesting.

It is important to maintain a trade-off between repeatability and variability, but the trend in new certification processes includes testing under real traffic conditions where repeatability cannot be guaranteed in any case. 

Finally, false positive testing is also necessary for each use case, i.e., an aborted cut-in or an emerging pedestrian who ultimately does not enter the crossing zone. Predictive systems must be able to predict both positive and negative cases and act accordingly.






\subsubsection{Adding real behaviors and interactions}
The only way to test real interactions between road users and automated vehicles in a safe manner is to have real road users in a simulated environment where they can safely interact with automated vehicles that will be equipped with adaptive motion planning and HMI strategies based on predictive features. The benefits of using simulated environments with real road users in the loop are two-fold: on the one hand, all variables and conditions can be fully controlled during the execution of the tests; on the other hand, the reactions of road users can be accurately measured, providing the means to assess the effect that the autonomous vehicle’s actions cause on other agents. This will definitely open the gate to the development of autonomous vehicles with real interacting capabilities that will mimic or even surpass human driving abilities.    




%


\section{Conclusion}
\label{sec:conclusions}

Physical tests conducted in controlled environments on test tracks are mainly developed to evaluate last-second reactions and do not allow changes in the trajectory or speed of the vehicle under test until the last moment, even if the vehicle has anticipated the situation correctly, which drastically reduces the ability of advanced predictive systems to improve safety and comfort. The conducted tests have proven that the developed advanced predictive systems have accomplished them increasing safety and comfort compared with basic systems even with these limitations.

Current testing procedures present some limitations to assess the real potential of advanced predictive systems in real driving situations. These limitations are totally opened scenarios, interactions limited to a single element, fixed and known appearance of the interacting element, strict repetitive conditions, execution of preprogrammed trajectories, rigid body movements, and the impossibility of anticipating. Vision-based systems may be more undervalued than radar and lidar-based systems because they perceive more limitations.

After analyzing the above limitations, the ability to measure the actual performance of predictive systems can be potentially increased by randomizing experiments and adding probabilistic noise to the dummies' linear trajectories. The development of more realistic dummies can increase reality in actions, such as pedestrians bending or turning their torso, cyclists turning their heads, or vehicles with robotized turn signals. Several interacting dummies need to be included in the tests to create more complex, but essentially the same scenarios. Tests need to be improved in terms of variability by increasing the number of possible final situations including different behaviors such as fully and partially developed, or even aborted maneuvers to correctly evaluate the goodness of the advanced predictive systems. In the same line, it is necessary to implement non-last-second tests to improve safety and comfort by anticipating oncoming events.

Our future work is mainly focused on addressing the implementation of the proposals discussed in this paper within the regulation and working groups of various institutions such as Euro NCAP (e.g., AEBS and AD, OSM, HMI), EC GSR2, and UNECE VMAV.

\section*{Acknowledgment}

This work was mainly funded by research grant BRAVE Project, H2020 (Contract \#723021) and partially funded by research grants S2018/EMT-4362 SEGVAUTO 4.0-CM (Community Reg. Madrid), DPI2017-90035-R and PID2020-114924RB-I00 (Spanish Min. of Science and Innovation). D. Fern\'{a}ndez Llorca acknowledges funding from the HUMAINT project by DG JRC of the European Commission.
We want to express our utmost gratitude to UTAC for providing the proving grounds and dummies and for the support in conducting the experiments.
\ifCLASSOPTIONcaptionsoff
  \newpage
\fi

\bibliographystyle{bibliography_ieee/IEEEtran}
\bibliography{bibliography_ieee/IEEEabrv,mybibfile}

%

\begin{IEEEbiography}[{\includegraphics[width=1in,height=1.25in,clip,keepaspectratio]{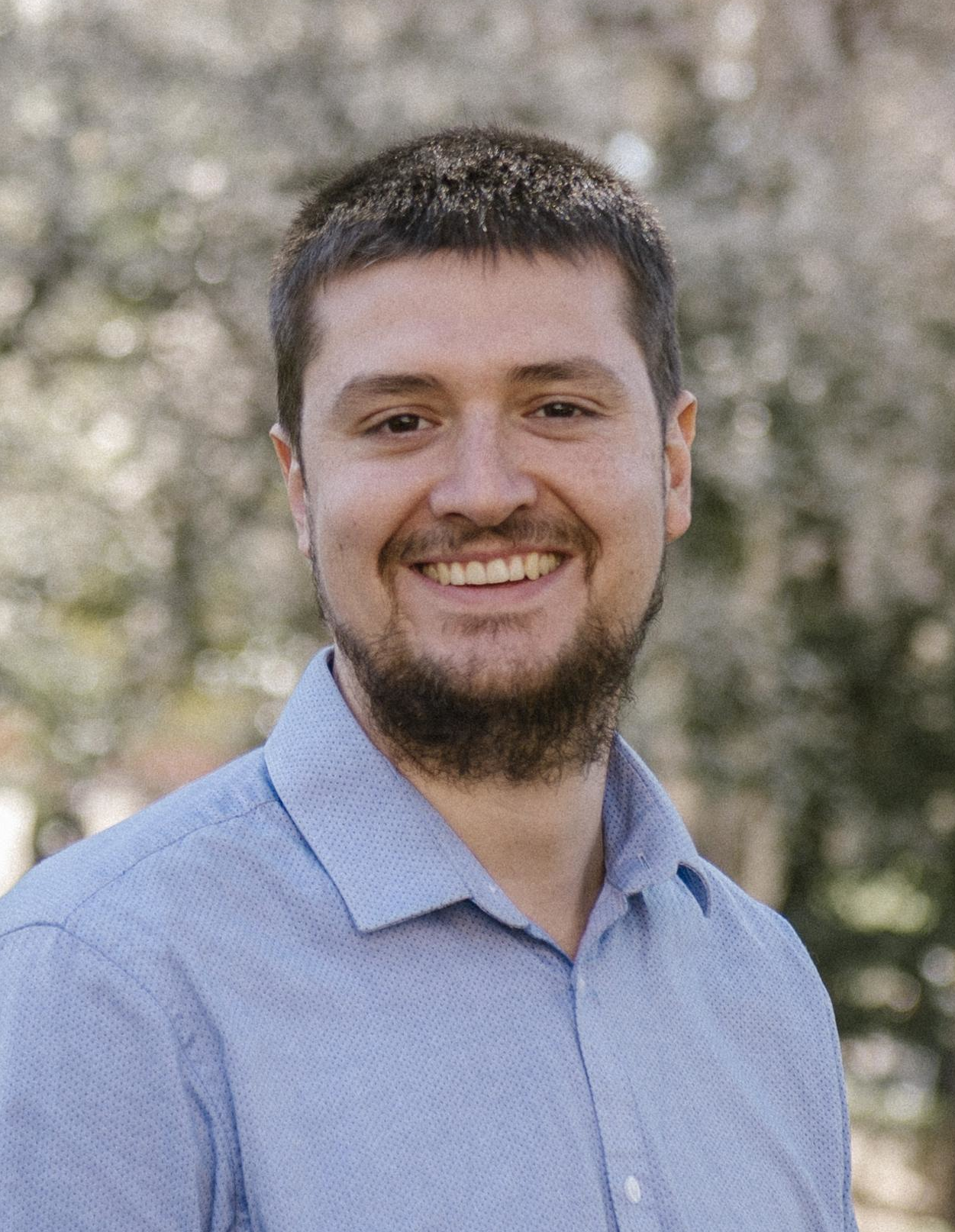}}]{Rub\'en Izquierdo Gonzalo} received the Bachelor's degree in electronics and industrial automation engineering in 2014, the M. S. in industrial engineering in 2016, and the Ph.D. degree in information and communication technologies 2020 from the Universidad de Alcalá (UAH). He is currently Assistant Professor at the Computer Engineering Department, UAH. His research interest is focused on the prediction of vehicle behaviors but also in control algorithms for highly automated and cooperative vehicles. His work has developed a predictive ACC and AES system for cut-in collision avoidance.
\end{IEEEbiography}

\begin{IEEEbiography}[{\includegraphics[width=1in,height=1.25in,clip,keepaspectratio]{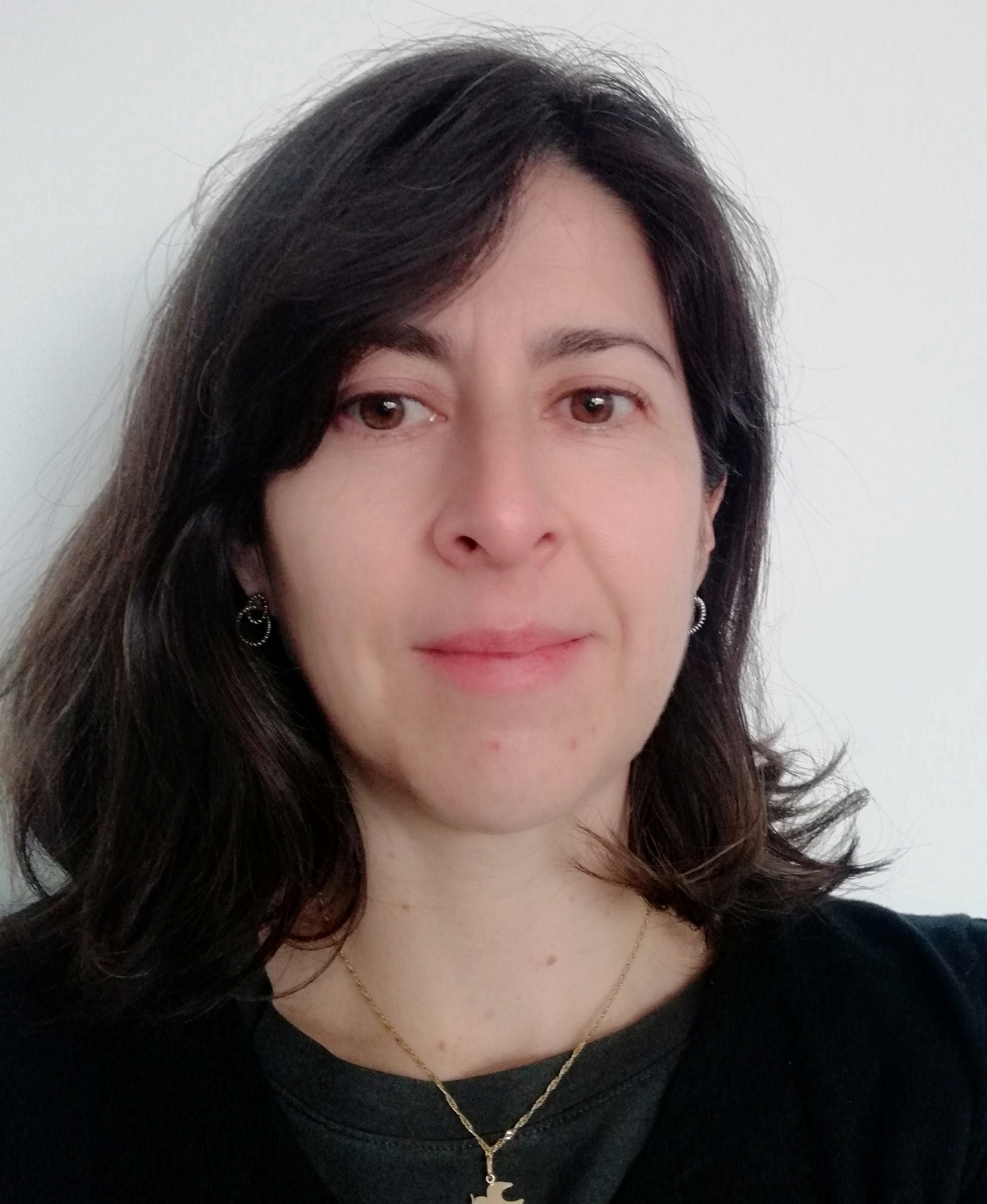}}]{Carlota Salinas Maldonado} received his B.Sc. in Industrial Engineering from the Universidad Privada Boliviana (UPB) in 2001 and the Ph.D. in Engineering and Automatics from the Universidad Complutense de Madrid (UCM) in 2015. She is currently an Assistant Professor at the Computer Engineering Department, University of Alcalá (UAH). Her research interests include autonomous vehicle navigation, data fusion systems, LiDAR, computer vision and machine learning algorithms.
\end{IEEEbiography}

\begin{IEEEbiography}[{\includegraphics[width=1in,height=1.25in,clip,keepaspectratio]{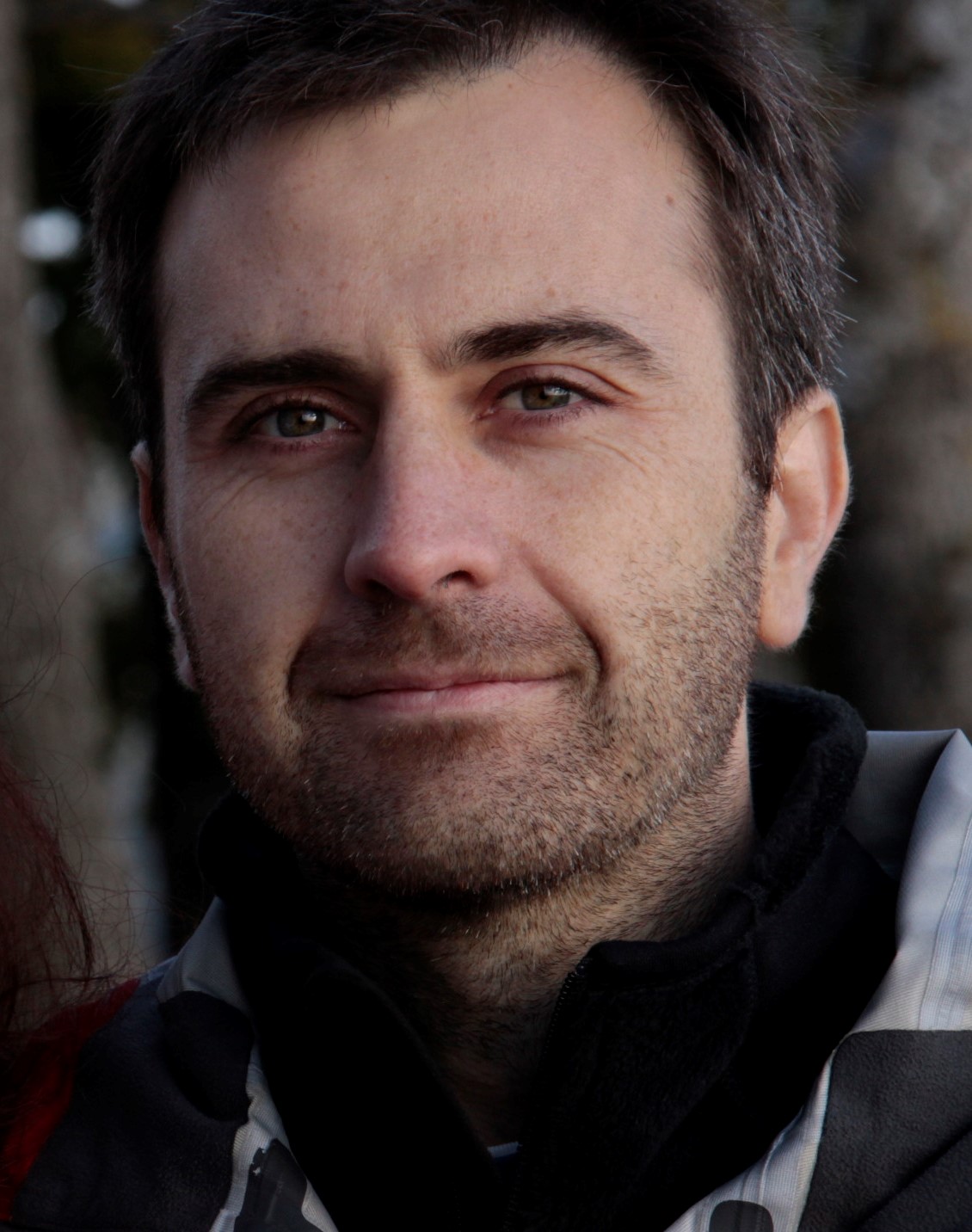}}]{Javier Alonso} received the degree in Computer Science at the Polytechnic University of Madrid (UPM) in 2001 and the PhD in 2009. He is currently Assistant Professor at the Computer Engineering Department, University of Alcalá (UAH). His research interests are focused in autonomous vehicle navigation and control. He was awarded with the finalist prize for the best PhD thesis in Intelligent Control 2010 by the Spanish Committee of Automatics. He succeeded in obtaining a Marie Curie Intra-European Fellowship to work on Cognitive Vehicles at the Karlsruher Institut für Technologie (KIT), Germany. And he has received the XIII Prize of the Social Council of the UAH to the University-Society knowledge transfer in 2018. 
\end{IEEEbiography}

\begin{IEEEbiography}[{\includegraphics[width=1in,height=1.25in,clip,keepaspectratio]{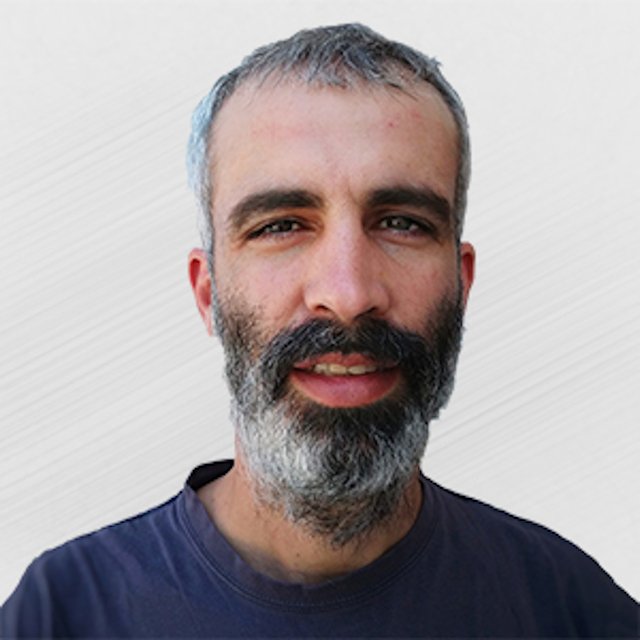}}]{Ignacio Parra Alonso} received the M.S. and Ph.D. degrees in telecommunications engineering from the University of Alcalá (UAH), in 2005 and 2010, respectively. He is currently an Associate Professor with the Computer Engineering Department, UAH. His research interests include intelligent transportation systems and computer vision. He received the Master Thesis Award in eSafety from the ADA Lectureship at the Technical University of Madrid, Spain, in 2006.
\end{IEEEbiography}

\begin{IEEEbiography}[{\includegraphics[width=1in,height=1.25in,clip,keepaspectratio]{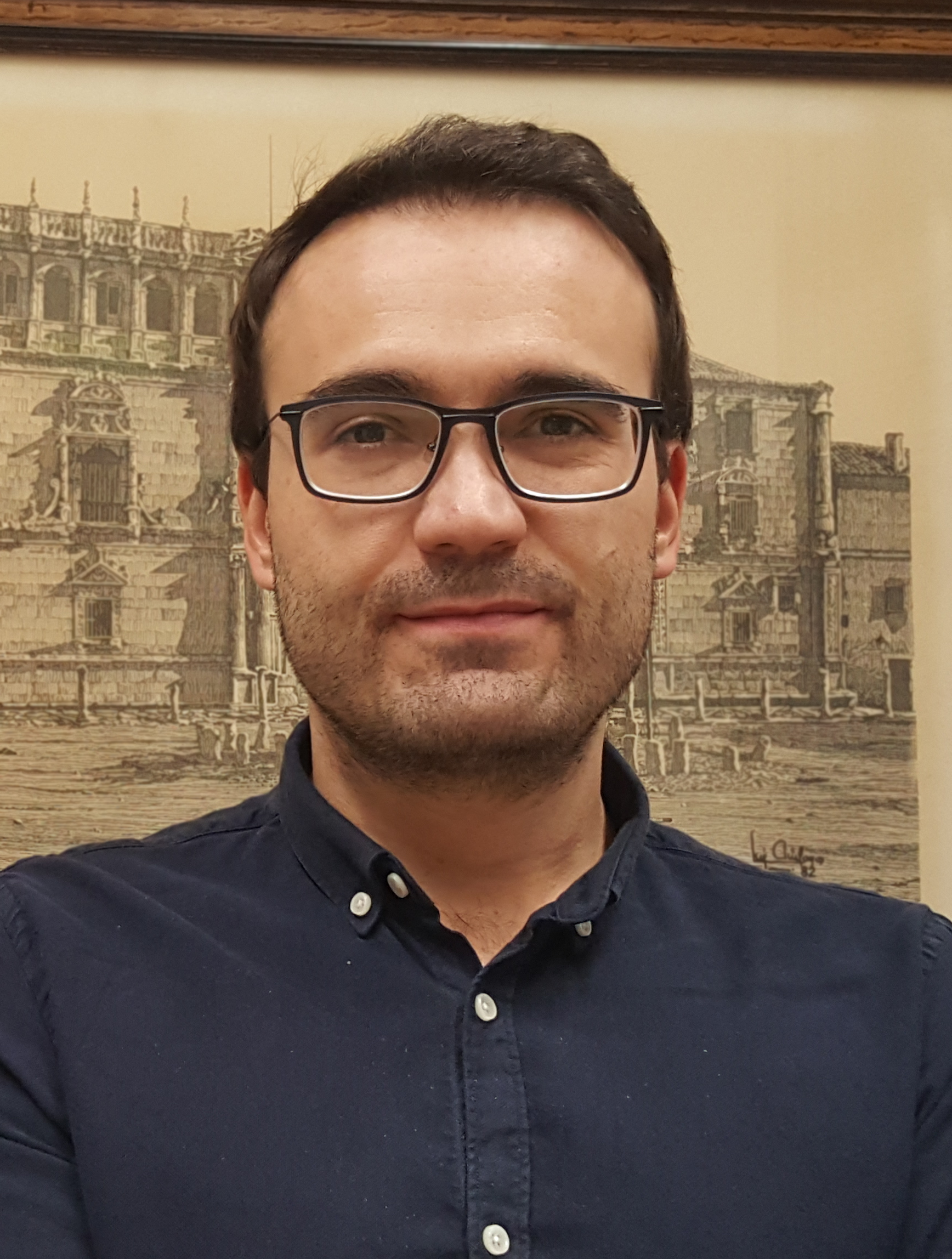}}]{David Fern\'andez-Llorca} (Senior Member, IEEE) received the Ph.D degree in telecommunication engineering from the University of Alcal\'a (UAH) in 2008. He is currently Scientific Officer at the European Commission - Joint Research Center. He is also Full Professor with UAH. He has authored over 130 publications and more than  10  patents. He received the IEEE ITSS Young Research Award in 2018 and the IEEE ITSS Outstanding Application Award in 2013. He is Editor-in-Chief of the IET Intelligent Transport Systems. His current research interest includes trustworthy AI for transportation, predictive perception for autonomous vehicles, human-vehicle interaction, end-user oriented autonomous vehicles and assistive intelligent transportation systems.
\end{IEEEbiography}

\begin{IEEEbiography}[{\includegraphics[width=1in,height=1.25in,clip,keepaspectratio]{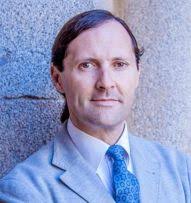}}]{Miguel \'Angel Sotelo} received the degree in Electrical Engineering in 1996 from the Technical University of Madrid, the Ph.D. degree in Electrical Engineering in 2001 from the University of Alcalá (Alcalá de Henares, Madrid), Spain, and the Master in Business Administration (MBA) from the European Business School in 2008. He is currently a Full Professor at the Department of Computer Engineering of the University of Alcalá. His research interests include Self-driving cars and Predictive Systems. He is author of more than 300 publications in journals, conferences, and book chapters. He has been recipient of the Best Research Award in the domain of Automotive and Vehicle Applications in Spain in 2002 and 2009, and the 3M Foundation Awards in the category of eSafety in 2004 and 2009. Miguel Ángel Sotelo has served as Project Evaluator, Rapporteur, and Reviewer for the European Commission in the field of ICT for Intelligent Vehicles and Cooperative Systems in FP6 and FP7. He is member of the IEEE ITSS Board of Governors and Executive Committee. Miguel Ángel Sotelo served as Editor-in-Chief of the Intelligent Transportation Systems Society Newsletter (2013), Editor-in-Chief of the IEEE Intelligent Transportation Systems Magazine (2014-2016), Associate Editor of IEEE Transactions on Intelligent Transportation Systems (2008-2014), member of the Steering Committee of the IEEE Transactions on Intelligent Vehicles (since 2015), and a member of the Editorial Board of The Open Transportation Journal (2006-2015). He has served as General Chair of the 2012 IEEE Intelligent Vehicles Symposium (IV’2012) that was held in Alcalá de Henares (Spain) in June 2012. He was recipient of the 2010 Outstanding Editorial Service Award for the IEEE Transactions on Intelligent Transportation Systems, the IEEE ITSS Outstanding Application Award in 2013, and the Prize to the Best Team with Full Automation in GCDC 2016. He was President of the IEEE Intelligent Transportation Systems Society (2018-2019).
\end{IEEEbiography}






\end{document}